\documentclass[10pt]{article} %
\usepackage[accepted]{rlc}

\usepackage{amssymb}            %
\usepackage{mathtools}          %
\usepackage{mathrsfs}           %
\usepackage{graphicx}           %
\usepackage[space]{grffile}     %
\usepackage{url}                %

\usepackage{graphicx}
\usepackage{booktabs}
\usepackage{adjustbox}
\usepackage{amsmath}
\usepackage{amssymb}
\usepackage{wrapfig}
\usepackage{algorithm}
\usepackage[indLines=false]{algpseudocodex}
\usepackage[font=small]{caption}
\usepackage{multirow}
\usepackage{tabularx}
\usepackage{tikz}
\usetikzlibrary{positioning,shapes}
\usepackage{varioref}
\usepackage{hyperref}
\usepackage{stfloats}

\usepackage{tikz}
\usetikzlibrary{positioning,angles,quotes,shapes,fit,calc,arrows.meta,patterns,patterns.meta,decorations.pathreplacing,calligraphy}

\newcommand{\po}{\phantom{0}}
\usepackage{pgfplots}
\pgfplotsset{compat=1.18,
	every axis/.append style={
		scale only axis,
		enlargelimits=false,
		ylabel near ticks,
		xlabel near ticks,
		title style={yshift=-8pt, font=\footnotesize},
		label style={font=\footnotesize},
		ylabel style={yshift=-3.5pt},
		xlabel style={yshift=2pt},
		tick label style={font=\tiny},
        yticklabel style={xshift=1.5pt},
        }
    }

\usepackage{enumitem} %

\labelformat{equation}{(#1)}

\newcommand{\E}{\mathbb{E}}

\newcommand{\lb}{\left [}
\newcommand{\rb}{\right ]}
\newcommand{\lp}{\left (}
\newcommand{\rp}{\right )}

\newcommand{\cblock}[1]{
	\hspace{-6pt}
	\begin{tikzpicture}
		[
		node/.style={square, minimum size=10mm, thick, line width=2pt},
		]
		\node[fill={#1!25}, draw={#1}] {}; %
	\end{tikzpicture}%
}

\newcommand{\mycline}[1]{
	\hspace{-6pt}
	\begin{tikzpicture}
		[
		node/.style={square, minimum size=10mm, thick, line width=2pt},
		]
        \draw[fill={#1}, draw={#1}]	(0,0) rectangle (0.03,0.4);	
	\end{tikzpicture}%
}

\newcommand{\cdashed}{
	\hspace{-6pt}
	\begin{tikzpicture}   
		\draw[fill=none, draw=none] (0,0) circle (3pt);
		\draw[color=black, ultra thick, densely dashed] (-0.2,0.00) -- (0.3,0.00);
	\end{tikzpicture}
	\hspace{-6pt}
}

\definecolor{mydarkblue}{rgb}{0,0.08,0.45} %
\definecolor{sb_blue}{RGB}{31, 119, 180}
\definecolor{sb_orange}{RGB}{255, 127, 14}
\definecolor{sb_green}{RGB}{44, 160, 44}
\definecolor{sb_red}{RGB}{214, 39, 40}
\definecolor{sb_purple}{RGB}{148, 103, 189}
\definecolor{sb_brown}{RGB}{140, 86, 75}
\definecolor{sb_pink}{RGB}{227, 119, 194}

\graphicspath{ {./figures/} }

\title{Imitation Learning from Observation\\ through Optimal Transport}

\author{Wei-Di Chang\\
    wchang@cim.mcgill.ca\\
    McGill University
    \And
    Scott Fujimoto\\
    McGill University
    \And
    David Meger\\
    McGill University
    \And
    Gregory Dudek\\
    McGill University
    }

\begin{document}

\maketitle

\begin{abstract}
Imitation Learning from Observation (ILfO) is a setting in which a learner tries to imitate the behavior of an expert, using only observational data and without the direct guidance of demonstrated actions. 
In this paper, we re-examine optimal transport for IL, in which a reward is generated based on the Wasserstein distance between the state trajectories of the learner and expert. We show that existing methods can be simplified to generate a reward function without requiring learned models or adversarial learning. Unlike many other state-of-the-art methods, our approach can be integrated with any RL algorithm and is amenable to ILfO. We demonstrate the effectiveness of this simple approach on a variety of continuous control tasks and find that it surpasses the state of the art in the IlfO setting, achieving expert-level performance across a range of evaluation domains even when observing only a single expert trajectory \textit{without} actions. 
\end{abstract}

\section{Introduction}

Imitation Learning~(IL) is a widely used and effective tool for teaching robots complex behaviors. Although Reinforcement Learning~(RL) has demonstrated success in learning motor skills from scratch in real-world systems~\citep{haarnoja2018applications, kalashnikov2018scalable}, Imitation Learning~(IL) remains a proven and practical way to learn behaviors from demonstrations, without the need for a hand-tuned and engineered reward signal required for RL. However, acquiring access to expert actions can be highly impractical. For example, robotic systems that are too challenging to teleoperate smoothly or in applications where the action spaces of the demonstrator and the imitator do not match, such as in Sim-to-Real problems \citep{desai2020imitation}. 

Imitation Learning from Observation (ILfO) eliminates the need for demonstrated actions by learning behaviors from sequences of expert states instead of requiring both expert states and actions. Similar to how humans learn new skills from watching others, ILfO algorithms learn from observational data alone. 
Consequently, this reduces the cost of data collection, making ILfO algorithms instrumental for deploying IL in complex real-world systems.  %

Moving to the observation-only space, however, introduces new challenges. While IL algorithms can learn by copying actions, 
ILfO algorithms require more exploration to succeed~\citep{kidambi2021mobile}, as they can only indirectly imitate the expert through observed outcomes. 
This emphasis on exploration creates a further challenge in that the states visited by the learner are more likely to be distant or non-overlapping with those of the expert.  Distant states are problematic for imitation via distribution matching \citep{ho2016generative, ghasemipour2020divergence, firl2020corl}, as the widely used KL divergence is ill-defined for non-overlapping distributions. While IL methods can circumvent this problem by accelerating early learning with behavior cloning, ILfO methods must deal with randomly initialized policies, which are unlikely to behave similarly to an expert demonstrator.

The field of optimal transport has garnered much attention in recent years, with theoretical and computational developments allowing it to evaluate distances between distributions defined on high-dimensional metric spaces~\citep{cuturi2013sinkhorn, bonneel2015sliced}. The Wasserstein distance, in particular, can compare non-overlapping distributions and quantify the spatial shift between the supports of the distributions. These properties make it a natural alternative to KL divergence-based objectives used by existing methods. Moreover, the Wasserstein distance can be computed without requiring separate models or learned components. This makes the Wasserstein distance more computationally efficient and conceptually simpler than other methods that rely on incremental adversarial signals learned via online interaction~\citep{ho2016generative, DACkostrikov, SIL}.

Prior work \citep{SIL, PWIL, durugkar2021adversarial} based on the Wasserstein distance for IL or ILfO  relies on numerous techniques, such as adversarial or learned components, or designed for sample-inefficient on-policy RL algorithms. Building on prior work~\citep{SIL}, we introduce a simpler approach that does not require adversarial components or on-policy learning. 
Our resulting approach, Observational Off-Policy Sinkhorn (OOPS), generates a reward function for \textit{any} RL algorithm, which minimizes the Wasserstein distance between expert and learner state trajectories. We benchmark OOPS against existing methods proposed to optimize the Wasserstein distance~\citep{SIL, PWIL}, as well as current state-of-the-art ILfO algorithms~\citep{ghasemipour2020divergence, opolozhu2020} on a variety of continuous control tasks. OOPS outperforms state-of-the-art methods for ILfO, achieving near-expert performance in every evaluated task with only a single trajectory without observing any actions. To facilitate reproducibility, all of our code is open-sourced%
\footnote{Link removed for anonymization. Code in supplementary material.}.

\section{Background}

\textbf{Setting.} Our task is formulated by an episodic finite-horizon MDP $(\mathcal{S}, \mathcal{A}, \mathcal{P}, r, p_0, T)$,  with state space $\mathcal{S}$, action space $\mathcal{A}$, transition dynamics $\mathcal{P}: \mathcal{S} \times \mathcal{A} \times \mathcal{S} \rightarrow [0,1]$, reward function $r: \mathcal{S} \times \mathcal{A} \rightarrow \mathbb{R}$, initial state distribution $p_0: \mathcal{S} \rightarrow [0,1]$, and $T$ the horizon. While the overarching objective is to maximize reward, in the Imitation Learning from Observation (ILfO) setting, the agent never observes the true reward. Instead, ILfO algorithms must use sequences of states (trajectories~$\tau$), generated by an unknown expert, to infer a reward signal or objective. We therefore only assume access to a dataset $D_E$ of $N$ state-only trajectories, $D_E = \{\tau_0, \tau_1,..., \tau_{N-1}\}$.

\textbf{Optimal Transport.} Optimal Transport (OT) seeks to compute a matching between the source and target measures while minimizing the %
transport cost~\citep{villani2009optimal}. In our work, we aim to minimize the distance between the distribution of trajectories defined by the learner and the expert.

Writing out trajectories in terms of their transitions $\tau = \{(s_0, s_1), (s_1, s_2),..., (s_{T-1}, s_{T})\}$, and viewing each transition as a datapoint, 
forms a discrete measure $\alpha$ over the state transition space $\mathcal{S} \times \mathcal{S}$, with weights $\textbf{a}$ and locations $(s_i, s_{i+1})_E \in \mathcal{S}\times \mathcal{S}$ for the expert:
$\alpha = \sum^T_{i=0} a_i \sigma_{(s_i, s_{i+1})_E}$
where $\sigma_{(s_i, s_{i+1})}$ is the Dirac delta function at position $(s_i, s_{i+1})$. Similarly for the learner, with weights $\textbf{b}$ and locations $(s_i, s_{i+1})_\pi$ for the learner, the trajectory rollout forms the measure $\beta = \sum^T_{i=0} b_i \sigma_{(s_i, s_{i+1})_\pi}$~\citep{peyre2019computational}.
In each trajectory, we consider each timestep as being equally important, and as such restrict the weight vectors $\mathbf{a}$ and $\mathbf{b}$ to the uniform weight vectors: $\sum_{i=0}^T a_i = 1, a_i = \frac{1}{T}\, \forall\, 0<i< T$, and $ \sum_{i=0}^T b_i = 1, b_i = \frac{1}{T}\, \forall\, 0<i< T$.

While the Monge formulation of OT enforces a one-to-one matching between %
measures, the Kantorovich formulation relaxes the OT problem by allowing each source point to split mass: the mass at any source point may be distributed across several locations \citep{villani2009optimal, peyre2019computational}. This provides the Wasserstein distance (or Kantorovich metric) over a distance metric $d$: 
\begin{equation}
    W_p(\alpha, \beta) := \lp \min_P \lp \sum_i^T \sum_j^T d(\alpha_i, \beta_j)^p P_{i,j} \rp \rp^{\frac{1}{p}}, 
\end{equation}
which uses a coupling matrix $P \in \mathbb{R}^{n \times m}_+$, where $P_{i, j}$ is the mass flowing from bin $i$ to bin $j$:
\begin{equation}
    P \in \mathbb{R}^{n \times m} \text{ such that } \sum_j P_{i,j} = \mathbf{a} \text{ and } \sum_i P_{i,j} = \mathbf{b}. 
\end{equation}
The optimal coupling $P$ between $\alpha$ and $\beta$ gives us the minimal cost transport plan between the measures defined by the trajectories $\tau_\pi$ and $\tau_E$. 

\textbf{Sinkhorn distance.} The Sinkhorn distance~$W_\text{Sk}$ is an entropy regularized version of the Wasserstein distance~\citep{cuturi2013sinkhorn}, for $W_1$, with $p=1$ this equals: 
\begin{align}
W_\text{Sk}(\tau_\pi, \tau_E) := \min_{\tilde P} \sum^T_{i=0} \sum^T_{j=0} d(\alpha_i, \beta_j) \tilde P_{i,j} - \lambda \mathcal{H}(\tilde P),%
\end{align}
where the entropy term $\mathcal{H}(\tilde P) := \sum_{i=0}^T \sum_{j=0}^T \tilde P_{ij} \log \tilde P_{ij}$. For any given value of $\lambda > 0$, the optimal coupling matrix $\tilde P$ for $W_\text{Sk}$ can be computed efficiently using the iterative Sinkhorn algorithm~\citep{sinkhorn1967diagonal}. At the cost of convergence speed, as $\lambda$ approaches 0, the Wasserstein distance is recovered, while increasing its value blurs out the transport matrix and spreads the mass between the two measures. This approximation is useful as it provides a computationally efficient method for estimating the optimal coupling matrix for the Wasserstein distance $\tilde P \approx P$ for small $\lambda$, where $W_\text{Sk}$ upper bounds $W_1$.

\section{Related Work}

\textbf{Imitation Learning.} Learning from Demonstrations (LfD) approaches can be generally classified into two types of approaches: IL methods, which learn directly from expert data, and Inverse Reinforcement Learning (IRL) methods~\citep{ziebart2008maximum} which infer a reward function that is optimized by RL. GAIL~\citep{ho2016generative} and related methods \citep{DACkostrikov, fu2017learning} leverage adversarial training. These methods optimize a distribution matching objective between the state-action distribution of the learner and the expert, in terms of various probability divergence metrics~\citep{ho2016generative, ghasemipour2020divergence, kostrikov2018discriminator, firl2020corl}. Each divergence objective leads to distinct imitative behavior (zero-forcing or mean-seeking or both), which can be exploited in different scenarios~\citep{ke2019imitation}. In contrast, our approach minimizes a Wasserstein distance-based objective, better suited for our ILfO context.

\textbf{Imitation Learning from Observations.}
Due to the challenging nature of ILfO, many methods rely on learning a model, via an inverse dynamics model used to infer the missing actions of the expert~\citep{torabi2018behavioral}, use objectives based on the transition dynamics of the expert~\citep{jaegle2021imitation, chang2022flow}, or simply model the entire MDP~\citep{kidambi2021mobile}. Adversarial methods have also been adapted from the IL context~\citep{sun2019provably, torabi2018generative}. Another common theme is \textit{f}-divergence minimization, \citep{firl2020corl} derive an approach based on the analytical gradients of \textit{f}-divergences and show that different variants (FKL, RKL, JS) can be achieved through their framework. %
OPOLO~\citep{opolozhu2020}, leverages off-policy learning on top of an inverse dynamics model and adversarial training. 
As opposed to existing methods, 
our approach leverages the Wasserstein distance to compute a non-adversarial and model-free reward for ILfO.

\textbf{Optimal Transport for Imitation Learning.} Minimization of the Wasserstein distance for IL has been previously considered in~\citep{xiao2019wasserstein, wdail} through Wasserstein Generative Adversarial Network (WGAN)-inspired approaches~\citep{wgan}. In an adversarial policy learning set up similarly to GAIL \citep{ho2016generative} and by restricting the discriminator to be a 1-Lipschitz function, these approaches can minimize the $W_1$ distance between the policy and the reference trajectory data distribution. However these methods suffer from the drawbacks of adversarial frameworks, which are hard to optimize and tune~\citep{arjovsky2017towards}, and have been shown to be poor estimators of $W_1$~\citep{stanczuk2021wasserstein}. 

More recent works~\citep{SIL, PWIL, haldar2023watch} use Wasserstein distance solvers, or related approximations, for IL. 
Our approach is closely based on
Sinkhorn Imitation Learning (SIL) \citep{SIL}, which uses the Sinkhorn distance~\citep{cuturi2013sinkhorn} to compute an entropy regularized Wasserstein distance between the state-action occupancy of the learner and expert. However, rather than use an upper bound defined by off-policy samples, they use on-policy RL~\citep{trpo} to optimize the cosine distance over the representation space of an adversarial discriminator trained alongside the imitation agent. In our work, we found that we can vastly improve sample efficiency by using an off-policy agent instead and can consider a more straightforward objective without adversarial or learned representations, an aspect previously thought required for good performance. Another related approach, PWIL~\citep{PWIL}, uses a greedy formulation of the Wasserstein distance and matches the current state-action pair $(s, a)$ to its closest counterpart in the expert demonstration dataset at every rollout step.  %
In our experimental analysis~(\autoref{fig:approx_comparison}), 
we show that our approximation via the Sinkhorn distance creates a tighter upper bound of the true Wasserstein distance and is crucial for consistent performance.
Contrary to SIL and PWIL, %
we focus on ILfO, giving new results and insights into the capabilities of OT in this context, and show that our approach matches or outperforms existing state-of-the-art methods.

\section{Wasserstein Imitation Learning from Observational Demonstrations} \label{sec:main}

In this section, we introduce our approach for minimizing the Wasserstein distance between expert trajectories and learner rollouts. To do so, we derive a reward function based on the distance between state transitions in pairs of trajectories.

\textbf{Deriving a reward from the Wasserstein distance.} With the absence of a true reward signal, the ILfO setting can be framed as a divergence-minimization problem, where the objective is to match the trajectory distributions of the learner and the expert. In our case, we choose the Wasserstein distance as a metric for this task. Unlike the widely used KL divergence, the Wasserstein distance is defined for distributions with non-overlapping support, making it amenable to scenarios where the behavior of the learner and the expert may be particularly distinct. %
We can define our ILfO task as minimizing the Wasserstein distance~$W_1$ between trajectories~$\tau_\pi$ sampled from the learner policy~$\pi$ and example trajectories~$\tau_E$ provided by an expert~$E$:
\begin{align}\label{eq:W_objective}
    \min_\pi \E_{\tau_\pi, \tau_E} \lb W_1(\tau_\pi, \tau_E) \rb = \min_\pi \E_{\tau_\pi, \tau_E} \lb \min_P \lp \sum_{i=0}^{T} \sum_{j=0}^{T} d((s_i, s_{i+1})_\pi, (s_j, s_{j+1})_E) P_{i,j} \rp \rb. %
\end{align} 
As the Wasserstein distance between a pair of trajectories can be defined as a sum over each of the transitions in each trajectory, for a given coupling matrix $P$, we can define a reward function 
\begin{align}\label{eq:wild_reward}
    \tilde r_t(s_t, s_{t+1}|\tau_\pi, \tau_E, P) := - \sum_{j=0}^{T} d((s_t, s_{t+1})_\pi, (s_j, s_{j+1})_E) P_{t,j}, %
\end{align}
such that summing the reward $\tilde r_t$ over a learner trajectory $\tau_\pi$ is equal to the Wasserstein distance
\begin{equation}\label{eq:wd_as_sum}
    W_1(\tau_\pi, \tau_E) = \min_P \lp -\sum_{i=0}^{T} \tilde r_t(s_t, s_{t+1}|\tau_\pi, \tau_E, P) \rp. 
\end{equation}
This naturally suggests an objective that involves the sum of rewards $\tilde r_t$ over learner trajectories
\begin{equation}
    J(\pi | E, P) := \E_{\pi, E} \lb \sum_{t=0}^T \tilde r_t(s_t, s_{t+1}|\tau_\pi, \tau_E, P) \rb,
\end{equation}
where our original objective (\autoref{eq:W_objective}) can be recovered: %
\begin{equation}\label{eq:full_J}
\max_\pi \min_P J(\pi | E, P) = \min_\pi \E_{\tau_\pi, \tau_E} \lb W_1(\tau_\pi, \tau_E) \rb.
\end{equation}
As the optimal coupling matrix $P$ can be approximated by the iterative Sinkhorn algorithm~\citep{sinkhorn1967diagonal}, the maximization of the objective $J$ %
with any RL algorithm, can be used as a replacement to minimizing the Wasserstein distance.

\textbf{Off-policy minimization of the Wasserstein distance.} As the reward $\tilde r_t(s_t, s_{t+1}|\tau_{\pi_n}, \tau_E, P)$ is defined as a function of a trajectory $\tau_{\pi_n}$ gathered by the learner $\pi_n$, any stale reward determined by trajectories from a previous policy $\pi_{n-m}$, $m \geq 1$, will not correspond with the Wasserstein distance of the current learner (as noted in \autoref{eq:wd_as_sum}). However, working with the assumption that a policy $\pi_{n}$ is better than any previous policy with respect to $J$, (i.e.\ $J(\pi_n) \geq J(\pi_{n-m})$ where $m \geq 1$), we remark that stale rewards provide an upper bound on the Wasserstein distance:
\begin{align}\label{eq:upper_bound}
    W_1(\tau_{\pi_n}, \tau_E) &= \min_P \lp -\sum_{i=0}^{T} \tilde r_t(s_t, s_{t+1}|\tau_{\pi_n}, \tau_E, P) \rp \leq \min_P \lp -\sum_{i=0}^{T} \tilde r_t(s_t, s_{t+1}|\tau_{\pi_{n-m}}, \tau_E, P) \rp. 
\end{align}
This means that previously collected off-policy trajectories can be used for learning in a principled manner, at the cost of the tightness of the upper bound of the Wasserstein distance. In our experimental results, we show that reusing prior data dramatically improves the sample efficiency of our algorithm over approaches which rely exclusively on online data~\citep{SIL}.

Our final approach, Observational Off-Policy Sinkhorn (OOPS) discovers a reward function in a similar manner to existing approaches~\citep{SIL, PWIL}, but in state transition space rather than state-action space. Unlike these prior approaches, OOPS avoids complexities such as adversarial learning or heuristic-based design of the reward function with multiple hyperparameters. OOPS is summarized in \autoref{alg:cap}.

\begingroup
\setlength{\textfloatsep}{0pt}
\begin{algorithm}[h]
\small
\caption{OOPS}\label{alg:cap}
\begin{algorithmic}[1]
\State {\bfseries Input:} Dataset of expert demonstrations~$D_E$.
\For{episodes $n = 1,..., N$}
\State Collect a trajectory from the environment. 
\State Compute the coupling matrix $P$ using the Sinkhorn algorithm~\citep{sinkhorn1967diagonal}. 
\State Compute the reward $\tilde r$ with $D_E$ and $P$ (\autoref{eq:wild_reward}). 
\State Train learner with a RL algorithm, and the collected trajectories and reward $\tilde r$. 
\EndFor
\end{algorithmic}
\end{algorithm}
\endgroup

\section{Experiments}
\subsection{Results}

\begin{figure*}[b]
\centering
\hfill
\begin{tikzpicture}[trim axis right, trim axis left]
\begin{axis}[
    width=0.175\textwidth,
    title={Hopper},
    ylabel={Total Reward (1k)},
    xlabel={Time steps (1M)},
    xtick={0, 0.2, 0.4, 0.6, 0.8, 1.0},
    xticklabels={0, 0.2, 0.4, 0.6, 0.8, 1.0},
    ytick={0, 500, 1000, 1500, 2000, 2500, 3000, 3500},
    yticklabels={0, 0.5, 1.0, 1.5, 2.0, 2.5, 3.0, 3.5},
]
\addplot graphics [
ymin=-200, ymax=3705,
xmin=-0.03, xmax=1.03,
]{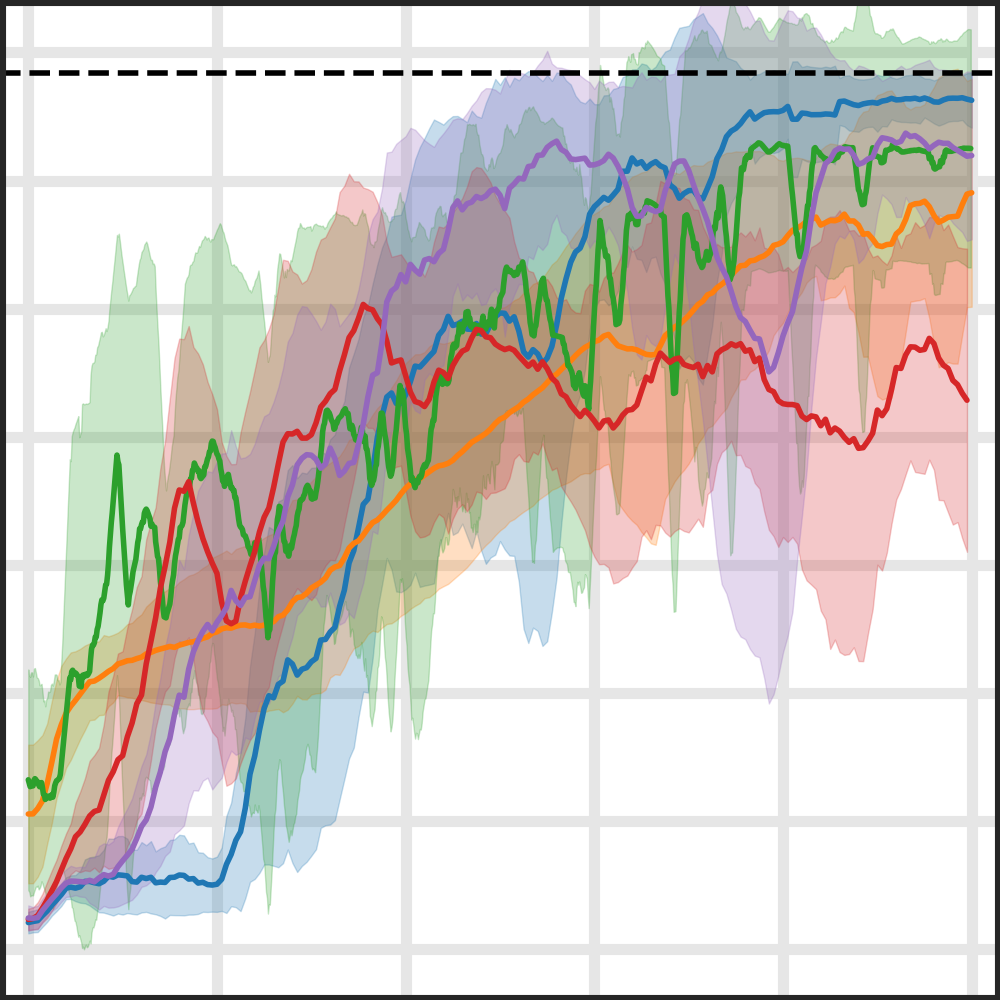};
\end{axis}
\end{tikzpicture}
\hspace{2pt}
\begin{tikzpicture}[trim axis right, trim axis left]
\begin{axis}[
    width=0.175\textwidth,
    title={\vphantom{p}Walker2d\vphantom{p}},
    xlabel={Time steps (1M)},
    xtick={0, 0.2, 0.4, 0.6, 0.8, 1.0},
    xticklabels={0, 0.2, 0.4, 0.6, 0.8, 1.0},
    ytick={0, 1000, 2000, 3000, 4000, 5000, 6000},
    yticklabels={0, 1, 2, 3, 4, 5, 6},
]
\addplot graphics [
ymin=-200, ymax=4700,
xmin=-0.03, xmax=1.03,
]{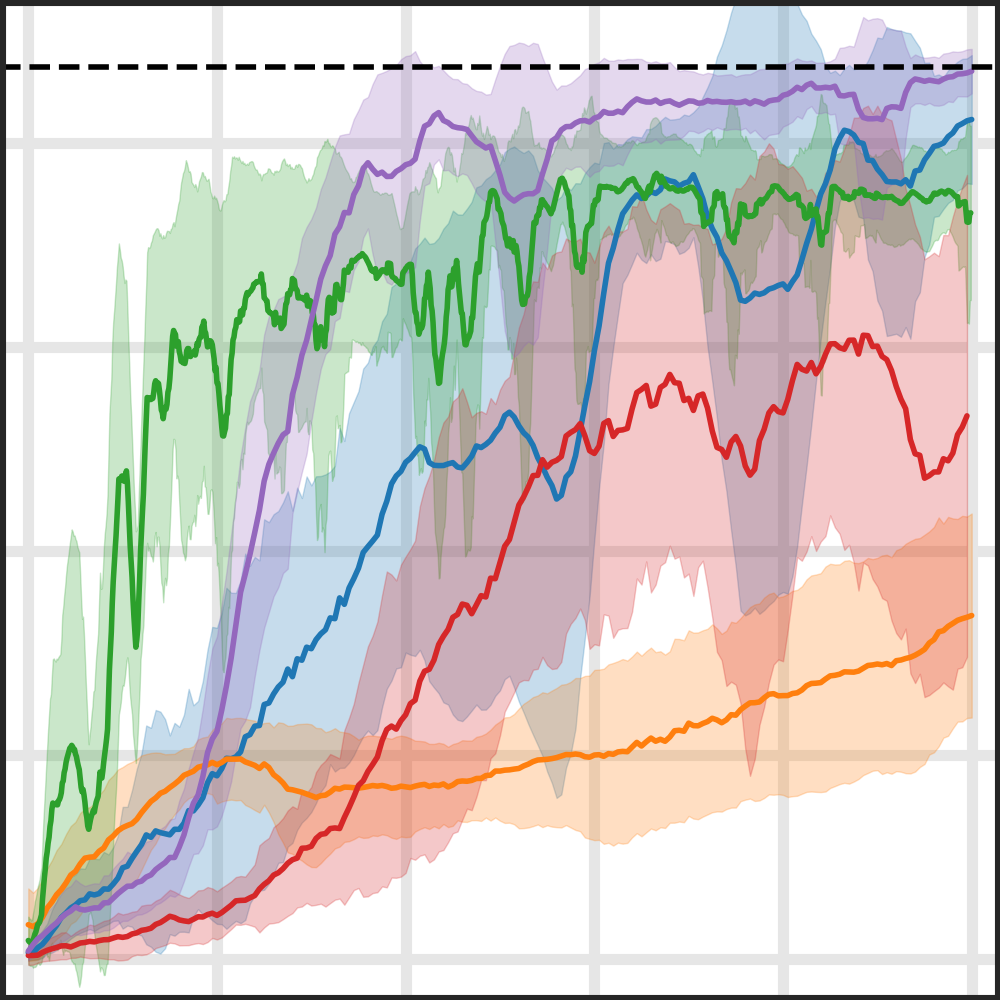};
\end{axis}
\end{tikzpicture}
\hspace{4pt}
\begin{tikzpicture}[trim axis right, trim axis left]
\begin{axis}[
    width=0.175\textwidth,
    title={\vphantom{p}HalfCheetah\vphantom{p}},
    xlabel={Time steps (1M)},
    xtick={0, 0.2, 0.4, 0.6, 0.8, 1.0},
    xticklabels={0, 0.2, 0.4, 0.6, 0.8, 1.0},
    ytick={0, 20, 40, 60, 80, 100, 120},
    yticklabels={0, 2, 4, 6, 8, 10, 12},
]
\addplot graphics [
ymin=-7, ymax=121.25,
xmin=-0.03, xmax=1.03,
]{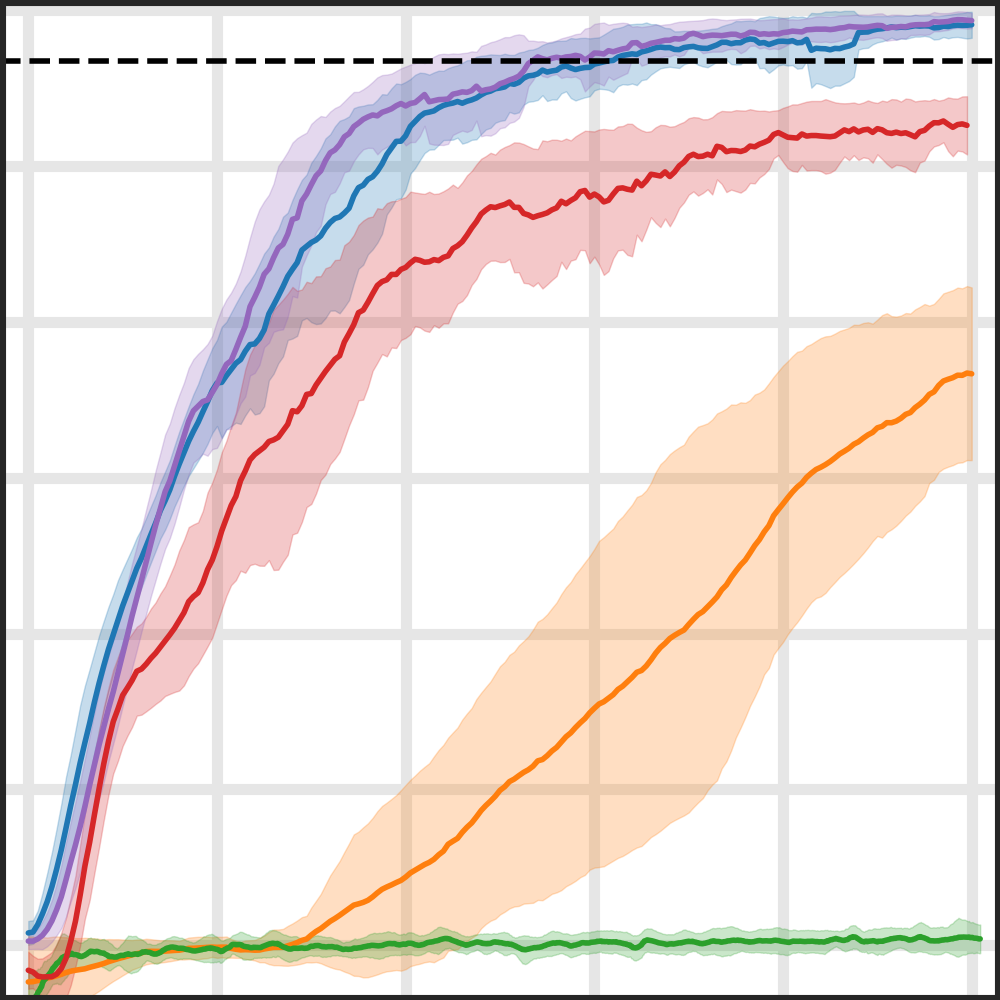};
\end{axis}
\end{tikzpicture}
\hspace{2pt}
\begin{tikzpicture}[trim axis right, trim axis left]
\begin{axis}[
    width=0.175\textwidth,
    title={\vphantom{p}Ant\vphantom{p}},
    xlabel={Time steps (1M)},
    xtick={0, 0.2, 0.4, 0.6, 0.8, 1.0},
    xticklabels={0, 0.2, 0.4, 0.6, 0.8, 1.0},
    ytick={-1000, 0, 1000, 2000, 3000, 4000, 5000, 6000},
    yticklabels={-1, 0, 1, 2, 3, 4, 5, 6},
]
\addplot graphics [
ymin=-1565, ymax=5400,
xmin=-0.03, xmax=1.03,
]{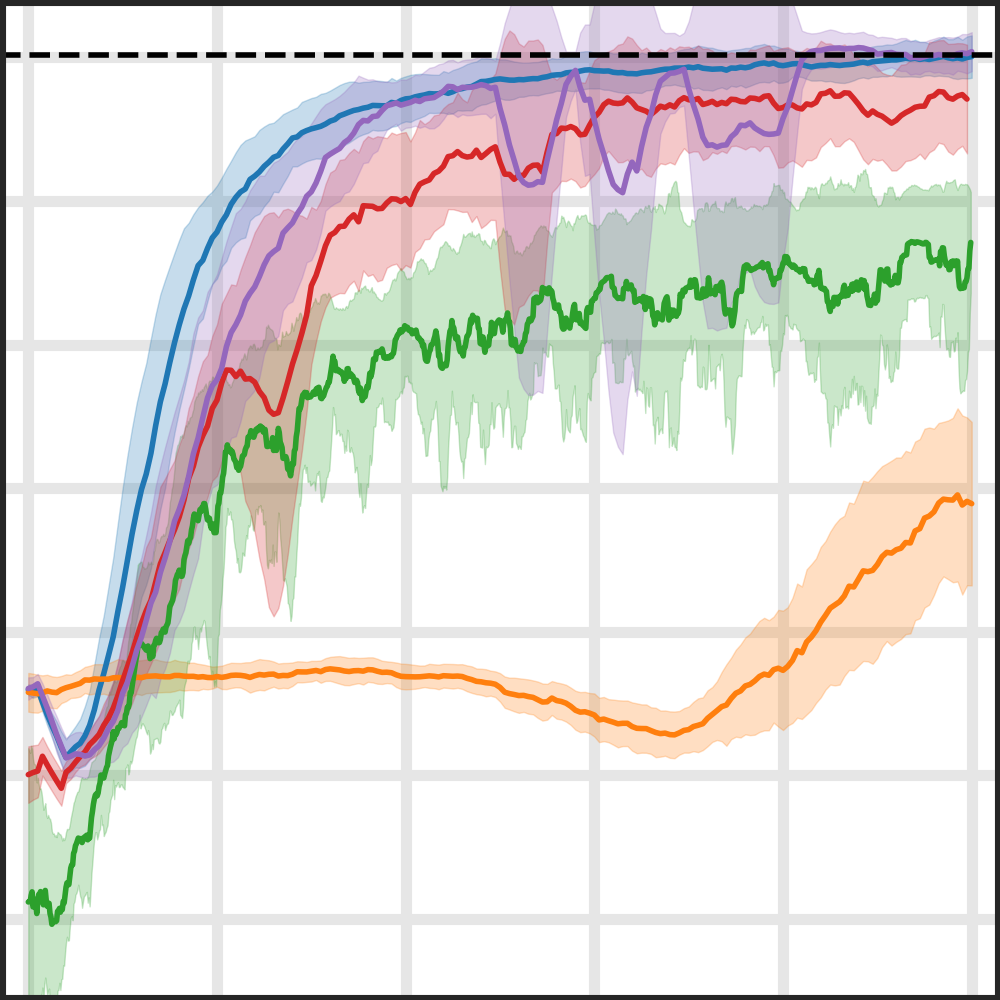};
\end{axis}
\end{tikzpicture}
\hspace{2pt}
\begin{tikzpicture}[trim axis right, trim axis left]
\begin{axis}[
    width=0.175\textwidth,
    title={\vphantom{p}Humanoid\vphantom{p}},
    xlabel={Time steps (1M)},
    xtick={0, 0.2, 0.4, 0.6, 0.8, 1.0},
    xticklabels={0, 0.2, 0.4, 0.6, 0.8, 1.0},
    ytick={0, 1000, 2000, 3000, 4000, 5000, 6000},
    yticklabels={0, 1, 2, 3, 4, 5, 6},
]
\addplot graphics [
ymin=-200, ymax=6300,
xmin=-0.03, xmax=1.03,
]{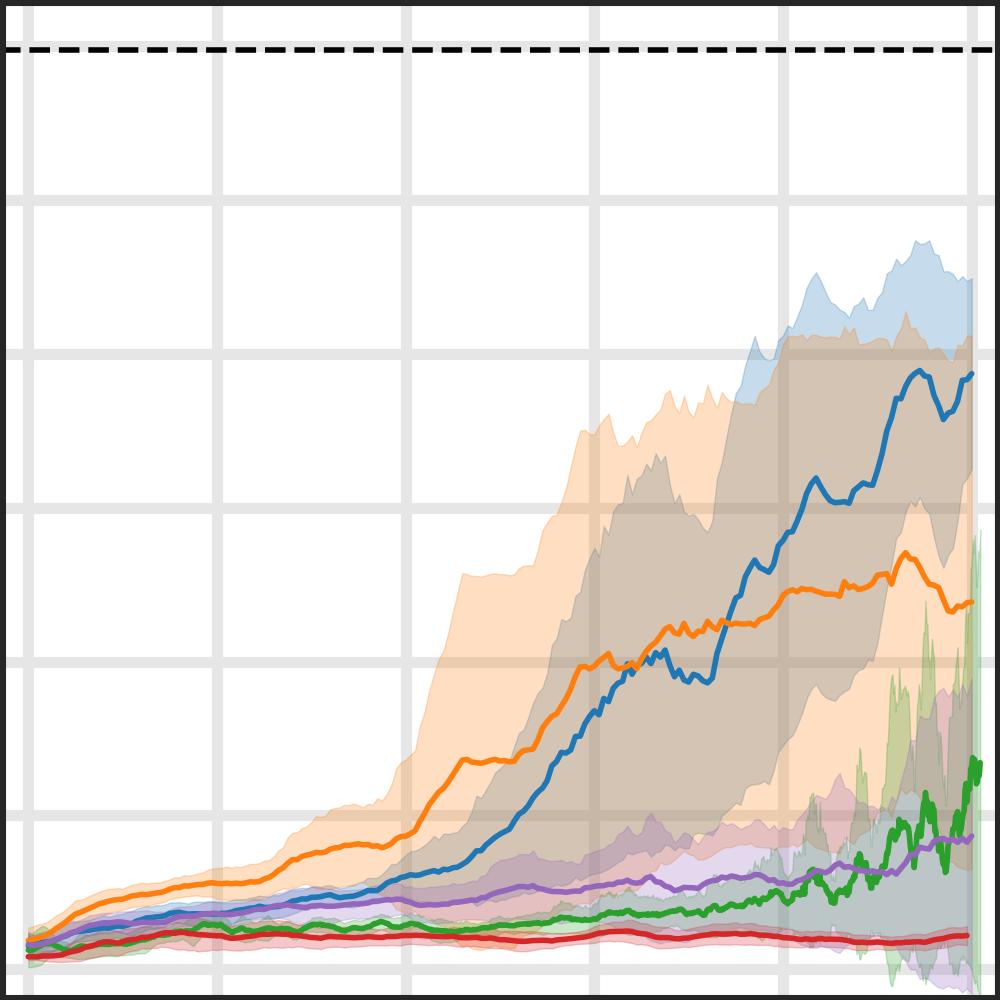};
\end{axis}
\end{tikzpicture}

\fcolorbox{gray}{white}{
\small \vphantom{$x^f$} \cblock{sb_blue} OOPS+TD3 \quad \cblock{sb_purple} OOPS+DDPG \quad \cblock{sb_orange} \textit{f}-IRL (FKL) \quad \cblock{sb_green} PWIL--$(s)$ \quad \cblock{sb_red} OPOLO \quad \cdashed Expert
}

\vspace{-4pt}
\caption{Learning curves for 1 expert demonstrations across 5 random seeds. The shaded area represents a standard deviation. OOPS+TD3 consistently matches or outperforms the baseline approaches.}\label{fig:learning_curves}
\vspace{-12pt}
\end{figure*}

We evaluate our algorithm on five MuJoCo locomotion benchmark environments from the OpenAI Gym suite~\citep{mujoco, OpenAIGym}, and three robotics tasks \citep{coumans2016pybullet, Tan-RSS-18} in the ILfO setting. For each environment, the dataset of expert trajectories~$D_E$ is generated via a pre-trained Soft Actor-Critic agent~\citep{haarnoja2018soft}. 

We use OOPS to generate a reward function for two RL algorithms, TD3~\citep{fujimoto2018addressing} and DDPG~\citep{DDPG}. Our baselines include state-of-the-art ILfO methods: f-IRL~\citep{firl2020corl} (its best-performing FKL variant in particular) and OPOLO~\citep{opolozhu2020}, as well as IL methods which also consider the Wasserstein distance: Primal Wasserstein Imitation Learning (PWIL)~\citep{PWIL} and Sinkhorn Imitation Learning (SIL)~\citep{SIL}.
In order to compare algorithms in the ILfO setting, we use the state-only version of PWIL, PWIL--$(s)$~\citep{PWIL}, and modify SIL \citep{SIL} %
by replacing the action $a$ in all pairs $(s, a)$ with the corresponding next state $s'$ in the transition. 
All algorithms are given a budget of 1M environment interactions (and 1M updates), are evaluated on 5 random seeds, and use the original implementations provided by the authors.

\begin{table*}[t]
{
\setlength{\tabcolsep}{4pt}
\centering
\footnotesize
\newcommand{\pp}{\phantom{.}}
\begin{tabularx}{0.99\textwidth}{cl*5{>{\raggedleft\arraybackslash}X}}
\toprule
{\# Expert} &  \multirow{2}{*}{Algorithm}                        & Hopper~~~         & Walker2d~~~       & HalfCheetah    & Ant~~~~~~            & Humanoid~~       \\
   {~Traj.} &                          &  3420  $\pm$ 36\pp\po  &     4370 $\pm$ 124\pp  & 11340 $\pm$ 95\pp\po   & 5018  $\pm$ 140\pp    &  5973 $\pm$ 17\pp\po 
   \\
\midrule
\multirow{6}{*}{1}   & \textit{f}-IRL (FKL) & 0.91  $\pm$ 0.03        & 0.42 $\pm$ 0.10       & 0.63 $\pm$ 0.13         & 0.47   $\pm$ 0.10      &  0.47 $\pm$ 0.32        \\
                     & OPOLO                & 0.73 $\pm$ 0.09         & 0.80 $\pm$ 0.14        & 0.88  $\pm$ 0.02       & 0.89  $\pm$ 0.04       & 0.04 $\pm$ 0.01         \\
                     & SIL -- $(s,s')$      & 0.17 $\pm$  0.06        & 0.07 $\pm$   0.02      & -0.17 $\pm$  0.09      & -0.41  $\pm$ 0.07      & 0.07 $\pm$   0.00       \\
                     & PWIL -- $(s)$        & 0.91 $\pm$ 0.14         & 0.71 $\pm$ 0.30       & 0.01   $\pm$  0.01      & 0.76     $\pm$ 0.05     & 0.14 $\pm$ 0.14 \\
                     & OOPS+DDPG (Ours)          & 0.90  $\pm$ 0.10 & \textbf{0.99 $\pm$ 0.03} & \textbf{1.05 $\pm$ 0.01} & \textbf{1.00 $\pm$ 0.02} & 0.16  $\pm$ 0.20        \\
                     & OOPS+TD3 (Ours)          & \textbf{0.98  $\pm$ 0.02} & \textbf{0.95 $\pm$ 0.09} & \textbf{1.05 $\pm$ 0.01} & \textbf{1.00 $\pm$ 0.03} & \textbf{0.74  $\pm$ 0.04}        \\
\midrule
\multirow{6}{*}{4}   & \textit{f}-IRL (FKL) & 0.92  $\pm$ 0.04          & 0.38  $\pm$ 0.12       & 0.69  $\pm$ 0.12       & 0.38   $\pm$ 0.07            &  0.51 $\pm$ 0.28        \\
                     & OPOLO                & 0.72  $\pm$ 0.15          & 0.91  $\pm$ 0.03       & 0.90  $\pm$ 0.02       & \textbf{1.02 $\pm$ 0.04 }    & 0.20 $\pm$ 0.12         \\
                     & SIL -- $(s,s')$      & 0.25  $\pm$   0.07         & 0.09  $\pm$  0.03     & -0.22  $\pm$  0.14     & -0.61  $\pm$  0.22           & 0.07   $\pm$ 0.01       \\
                     & PWIL -- $(s)$        & \textbf{0.98 $\pm$ 0.02}   & 0.88  $\pm$ 0.03      & 0.00  $\pm$   0.02      & 0.78    $\pm$ 0.03           &  0.23 $\pm$ 0.28          \\
                     & OOPS+DDPG (Ours)          & 0.75  $\pm$ 0.34          & \textbf{0.96 $\pm$ 0.03} & \textbf{1.05 $\pm$ 0.01} & \textbf{0.99 $\pm$ 0.01} &  0.07 $\pm$ 0.01 \\
                     & OOPS+TD3 (Ours)          & \textbf{0.94  $\pm$ 0.07}          & \textbf{0.97 $\pm$ 0.01} & \textbf{1.05 $\pm$ 0.01} & \textbf{0.99 $\pm$ 0.03} &  \textbf{0.65 $\pm$ 0.15} \\
\midrule
\multirow{6}{*}{10}  & \textit{f}-IRL (FKL) & 0.91  $\pm$ 0.05            & 0.39  $\pm$ 0.09           & 0.65  $\pm$  0.10        & 0.39  $\pm$ 0.17       &  0.40  $\pm$ 0.22        \\
                     & OPOLO                &  0.66   $\pm$ 0.08        & \textbf{0.96 $\pm$ 0.04}   & 0.95   $\pm$ 0.01        & \textbf{1.00 $\pm$ 0.03 } & 0.16 $\pm$ 0.06        \\
                     & SIL -- $(s,s')$      & 0.17   $\pm$  0.09        & 0.08   $\pm$ 0.03          & -0.20    $\pm$  0.09     & -0.24  $\pm$ 0.11        &  0.07  $\pm$  0.00        \\
                     & PWIL -- $(s)$        & \textbf{0.98 $\pm$ 0.01} & 0.87  $\pm$ 0.08            & 0.01    $\pm$ 0.02       & 0.78   $\pm$ 0.04        & 0.23 $\pm$ 0.28  \\
                     & OOPS+DDPG (Ours)          & \textbf{0.93 $\pm$ 0.03} & 0.78 $\pm$ 0.39    & \textbf{1.03 $\pm$ 0.04} & 0.79 $\pm$ 0.38 & 0.21 $\pm$ 0.25\\
                     & OOPS+TD3 (Ours)          & \textbf{0.97 $\pm$ 0.01} & \textbf{0.95 $\pm$ 0.03}    & \textbf{1.05 $\pm$ 0.01} & \textbf{1.00 $\pm$ 0.02} & \textbf{0.64 $\pm$ 0.22}\\
\bottomrule

\end{tabularx}
\vspace{-4pt}
 \caption{Final performance of different ILfO algorithms at 1M timesteps, using 1, 4, 10 expert demonstrations. Values for each task are normalized by the average return of the expert. $\pm$ captures the standard deviation. The highest value and any within $0.05$ are \textbf{bolded}. The average un-normalized return of the expert is listed below each task. All results are averaged across 5 seeds and 10 evaluations.} \label{table:results}
\vspace{-15pt}
}
\end{table*}

\textbf{Locomotion.} We report the evaluation results of our approach compared against the four baseline algorithms in \autoref{table:results}, varying the number of expert demonstrations used for imitation. The learning curves for the single demonstration setting are shown in \autoref{fig:learning_curves}.

\begin{wraptable}{r}{0.6\textwidth}
\vspace{-8pt}
{\setlength{\tabcolsep}{2pt}
\centering
\footnotesize
\begin{tabularx}{0.6\textwidth}{cl*5{>{\raggedleft\arraybackslash}X}}
\toprule
\multicolumn{2}{l}{\multirow{2}{*}{\# Expert Traj.\ }}                         & BipedalWalker         &  Minitaur~~~~     & MinitaurDuck       \\
&                        &  318.90 $\pm$ 9.20  & 12.36 $\pm$ 0.75 & 10.68 $\pm$ 1.20   \\
\midrule
\multirow{3.00}{*}{~~~~1~~~~}  & OPOLO              &     0.96 $\pm$ 0.01      &   0.76  $\pm$ 0.08        &   \textbf{ 1.00 $\pm$ 0.04 } \\
                     & PWIL -- $(s)$     &   0.89 $\pm$ 0.01                   &    0.53 $\pm$ 0.19         &   0.30 $\pm$ 0.14 \\
                     & OOPS+TD3       & \textbf{0.93 $\pm$ 0.01}         & \textbf{1.01 $\pm$ 0.04 }   &  \textbf{0.94 $\pm$ 0.18} \\
\midrule
\multirow{3.00}{*}{4}  & OPOLO              &     \textbf{ 0.96 $\pm$ 0.01 }    &   0.84 $\pm$ 0.09    &   \textbf{ 1.01 $\pm$ 0.03 } \\
                     & PWIL -- $(s)$      &    0.90 $\pm$ 0.01               &    0.52 $\pm$ 0.15           &    0.21 $\pm$ 0.09  \\
                     & OOPS+TD3        & \textbf{0.92 $\pm$ 0.01}       & \textbf{0.91 $\pm$ 0.09}         &  \textbf{1.02 $\pm$ 0.05}  \\
\midrule
\multirow{3.00}{*}{10}  & OPOLO               &   \textbf{   0.98 $\pm$ 0.00 }   &   \textbf{ 0.98 $\pm$ 0.04  }   &    \textbf{ 1.00 $\pm$ 0.02  } \\
                     & PWIL -- $(s)$       &    0.88 $\pm$ 0.01               &    0.58 $\pm$ 0.09          &   0.15 $\pm$ 0.16 \\
                     & OOPS+TD3         & \textbf{0.93 $\pm$ 0.01}        & \textbf{1.03 $\pm$ 0.03}   &  \textbf{0.99 $\pm$ 0.09} \\
\bottomrule
\end{tabularx}
 \caption{Final performance of ILfO algorithms when using 1, 4, and 10 expert demonstrations. Values for each task 
 are normalized by the average return of the expert. $\pm$ captures the standard deviation. The highest value and any within $0.05$ are \textbf{bolded}. The average un-normalized return of the expert is listed below each task. Results are averaged across 5 seeds and 10 evaluations.}
 \label{table:new_envs}
\vspace{-14pt}
}
\end{wraptable}

OOPS+TD3 consistently matches or outperforms all baseline methods regardless of task and number of expert demonstrations. OOPS+DDPG roughly matches the performance of the expert in every environment other than Humanoid. The poor results on Humanoid are unsurprising, as previous results have demonstrated that DDPG tends to fail at the Humanoid task in the standard RL 
setting~\citep{haarnoja2018soft}. 
Regardless, since DDPG is known to underperform TD3 and SAC, matching the performance of the SAC expert suggests that the OOPS reward function can produce a stronger learning signal than the original task reward. %
This result indicates that OOPS might not be dependent on the choice of the RL algorithm, assuming the RL algorithm is capable of solving the desired task.

\textbf{Additional environments.} For the top three performing algorithms (OPOLO, PWIL--$(s)$, and OOPS+TD3), we benchmark on three additional robotic-centric tasks in \autoref{table:new_envs}.  
While OOPS+TD3 and OPOLO achieve a similar high performance when using all 10 expert demonstrations, OOPS+TD3 surpasses OPOLO when using fewer demonstrations.

\begin{figure*}[t]
\centering
\hfill
\begin{tikzpicture}[trim axis right, trim axis left]
\begin{axis}[
    width=0.175\textwidth,
    title={Hopper},
    ylabel={Total Reward (1k)},
    xlabel={Total Proxy Reward},
    xtick={0, -10, -20, -30, -40, -50},
    xticklabels={0, 10, 20, 30, 40, 50},
    ytick={0, 500, 1000, 1500, 2000, 2500, 3000, 3500},
    yticklabels={0, 0.5, 1.0, 1.5, 2.0, 2.5, 3.0, 3.5},
]
\addplot graphics [
ymin=-174.62, ymax=3787.98,
xmin=-50, xmax=3.76,
]{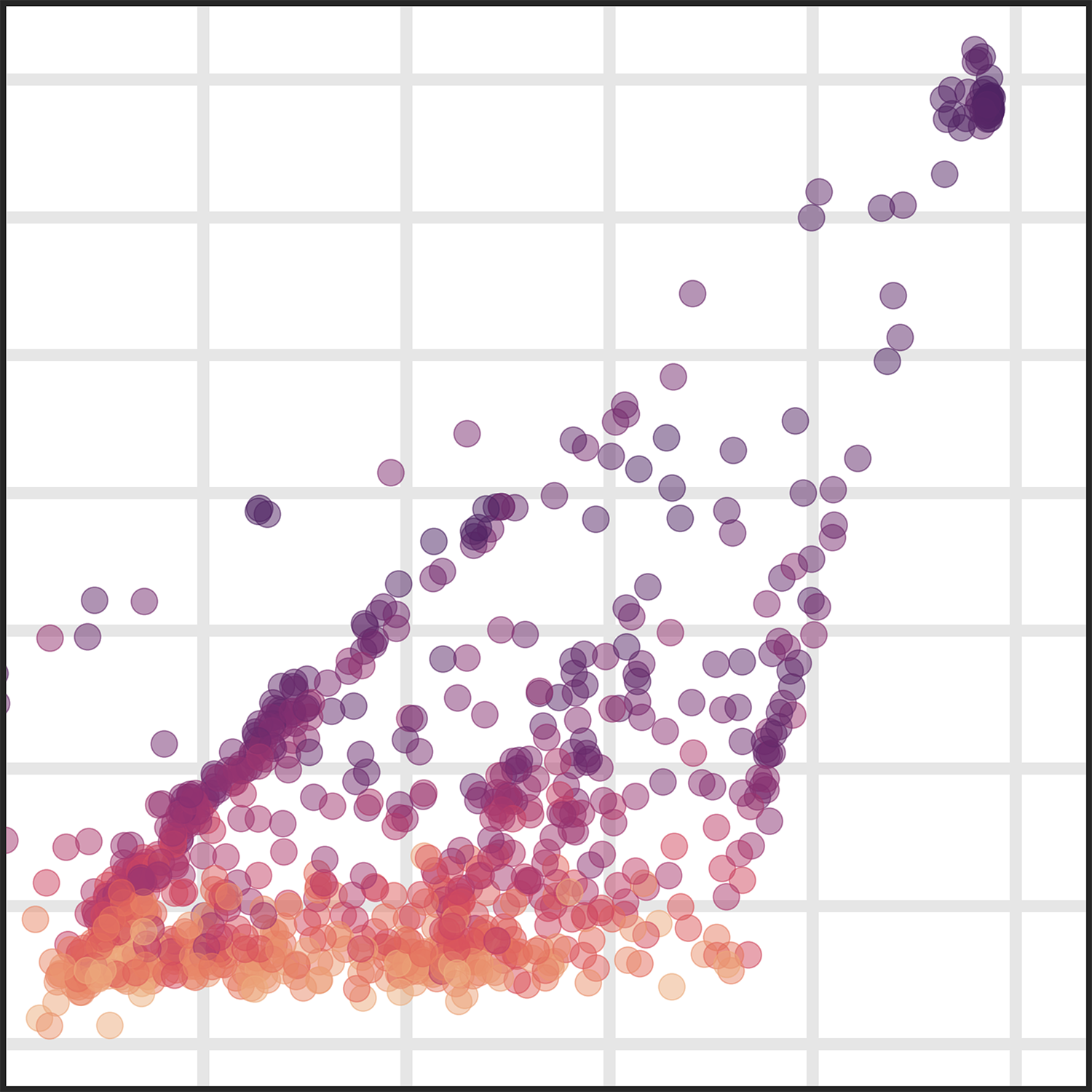};
\end{axis}
\end{tikzpicture}
\hspace{2pt}
\begin{tikzpicture}[trim axis right, trim axis left]
\begin{axis}[
    width=0.175\textwidth,
    title={\vphantom{p}Walker2d\vphantom{p}},
    xlabel={Total Proxy Reward},
    xtick={0, -10, -20, -30, -40, -50},
    xticklabels={0, 10, 20, 30, 40, 50},
    ytick={0, 1000, 2000, 3000, 4000, 5000, 6000},
    yticklabels={0, 1, 2, 3, 4, 5, 6},
]
\addplot graphics [
ymin=-186.64, ymax=4873.45,
xmin=-50, xmax=1,
]{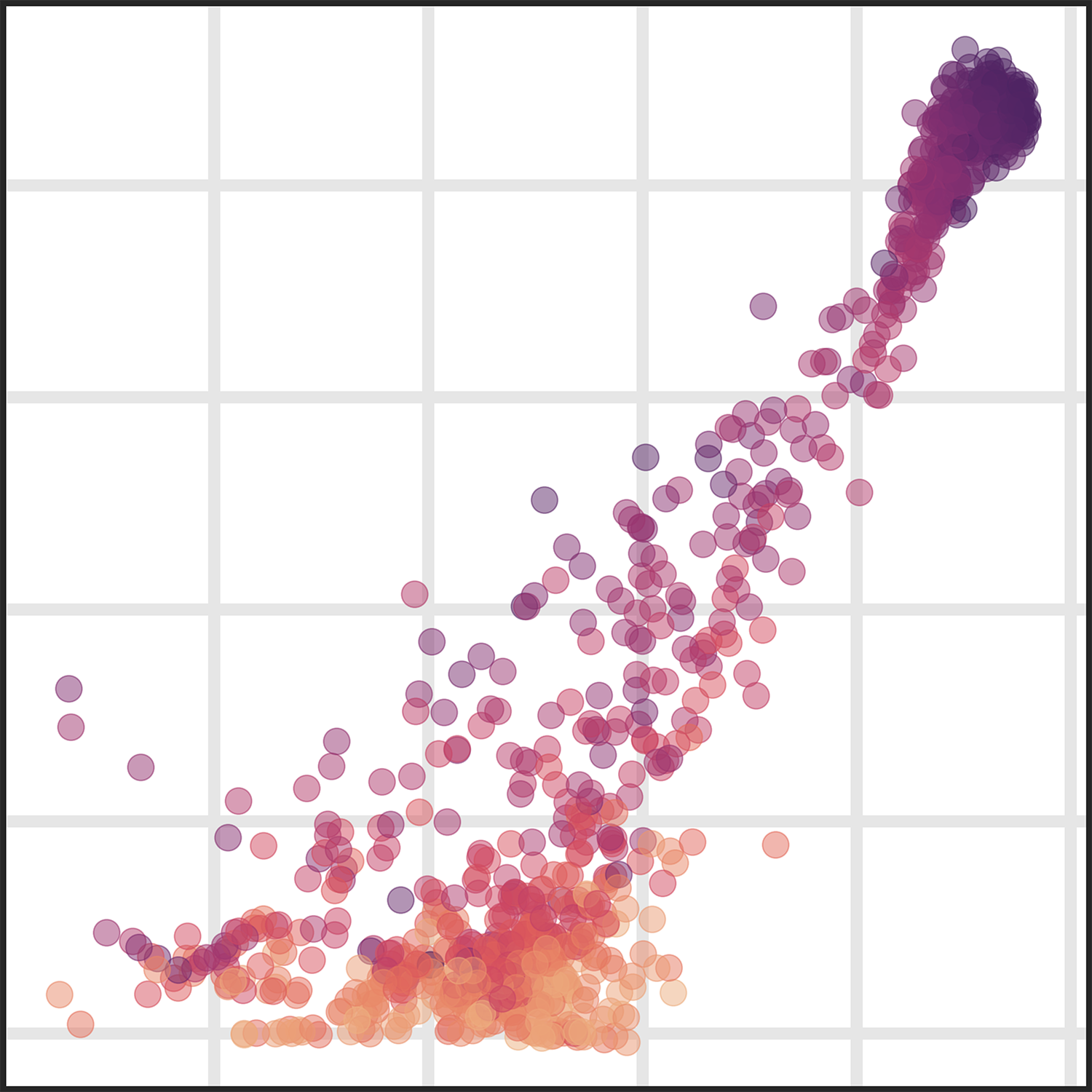};
\end{axis}
\end{tikzpicture}
\hspace{4pt}
\begin{tikzpicture}[trim axis right, trim axis left]
\begin{axis}[
    width=0.175\textwidth,
    title={\vphantom{p}HalfCheetah\vphantom{p}},
    xlabel={Total Proxy Reward},
    xtick={0, -2, -4, -6, -8, -10, -12, -14, -16},
    xticklabels={0, 2, 4, 6, 8, 10, 12, 14, 16},
    ytick={0, 20, 40, 60, 80, 100, 120},
    yticklabels={0, 2, 4, 6, 8, 10, 12},
]
\addplot graphics [
ymin=-10.19, ymax=120.71,
xmin=-15.6, xmax=0.6,
]{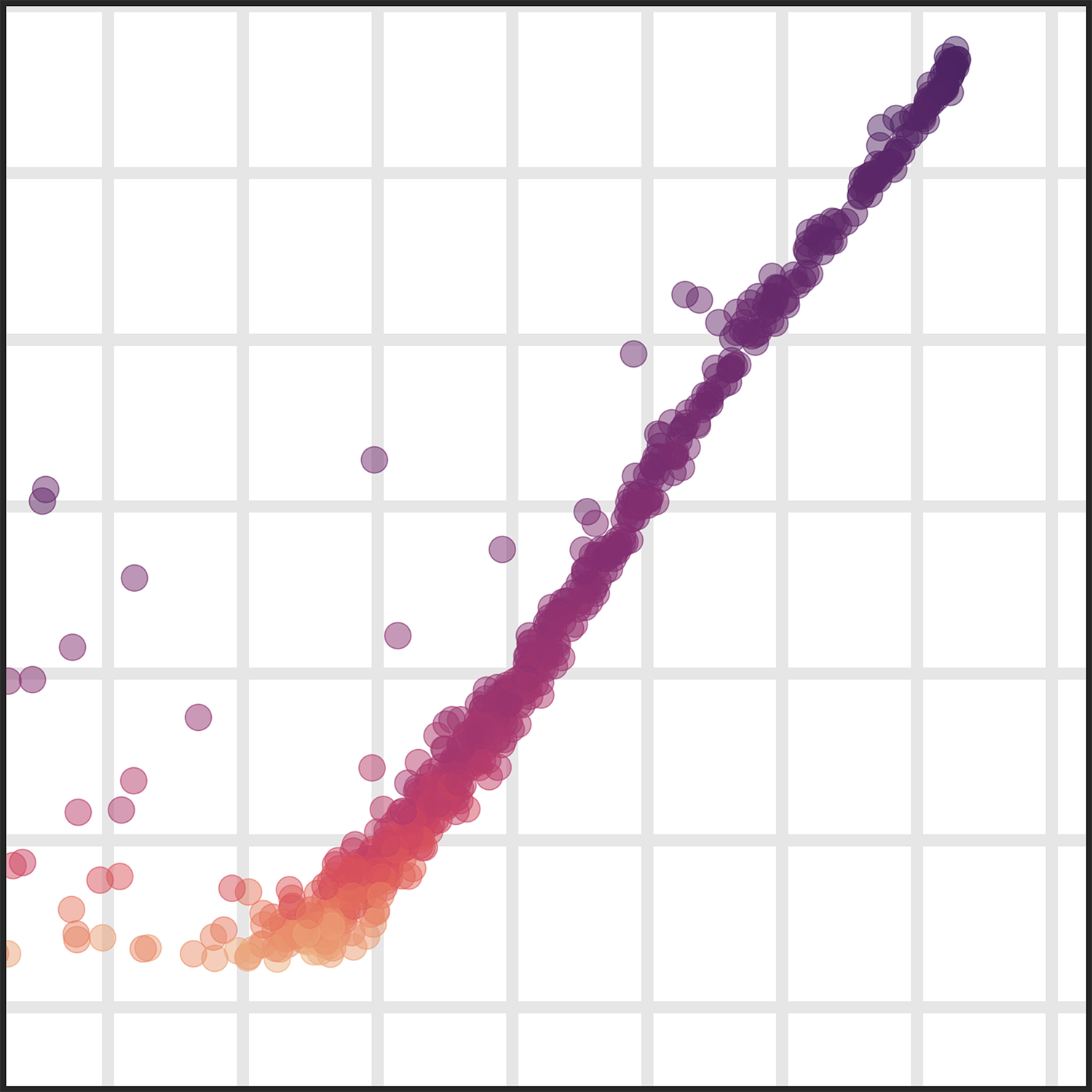};
\end{axis}
\end{tikzpicture}
\hspace{2pt}
\begin{tikzpicture}[trim axis right, trim axis left]
\begin{axis}[
    width=0.175\textwidth,
    title={\vphantom{p}Ant\vphantom{p}},
    xlabel={Total Proxy Reward},
    xtick={0, -10, -20, -30, -40, -50},
    xticklabels={0, 10, 20, 30, 40, 50},
    ytick={-1000, 0, 1000, 2000, 3000, 4000, 5000, 6000},
    yticklabels={-1, 0, 1, 2, 3, 4, 5, 6},
]
\addplot graphics [
ymin=-2112.83, ymax=5513.71,
xmin=-44.24, xmax=-2.23,
]{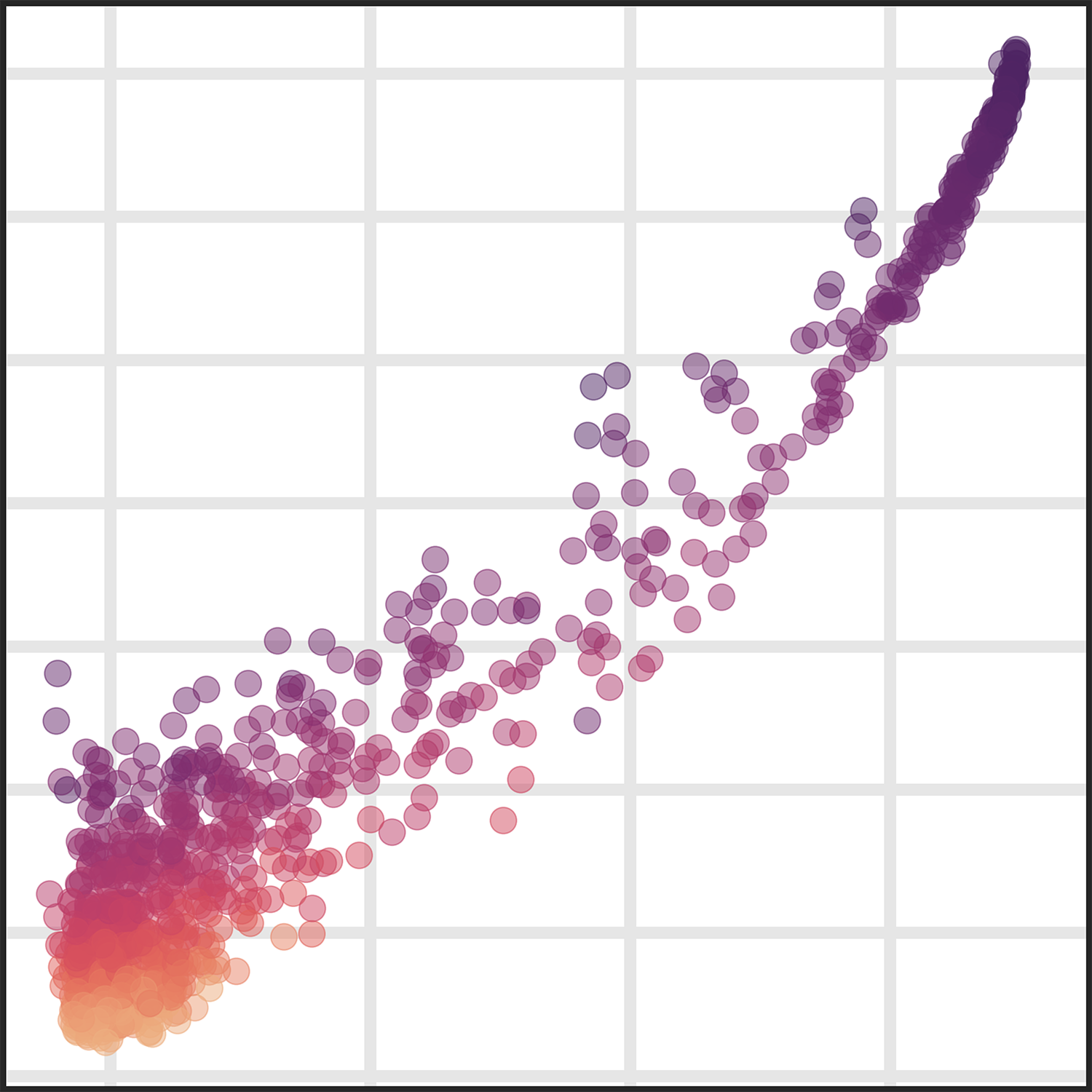};
\end{axis}
\end{tikzpicture}
\hspace{2pt}
\begin{tikzpicture}[trim axis right, trim axis left]
\begin{axis}[
    width=0.175\textwidth,
    title={\vphantom{p}Humanoid\vphantom{p}},
    xlabel={Total Proxy Reward},
    xtick={0, -20, -40, -60, -80, -100, -120},
    xticklabels={0, 20, 40, 60, 80, 100, 120},
    ytick={0, 1000, 2000, 3000, 4000, 5000, 6000},
    yticklabels={0, 1, 2, 3, 4, 5, 6},
]
\addplot graphics [
ymin=-186.64, ymax=5569.58,
xmin=-126, xmax=-5,
]{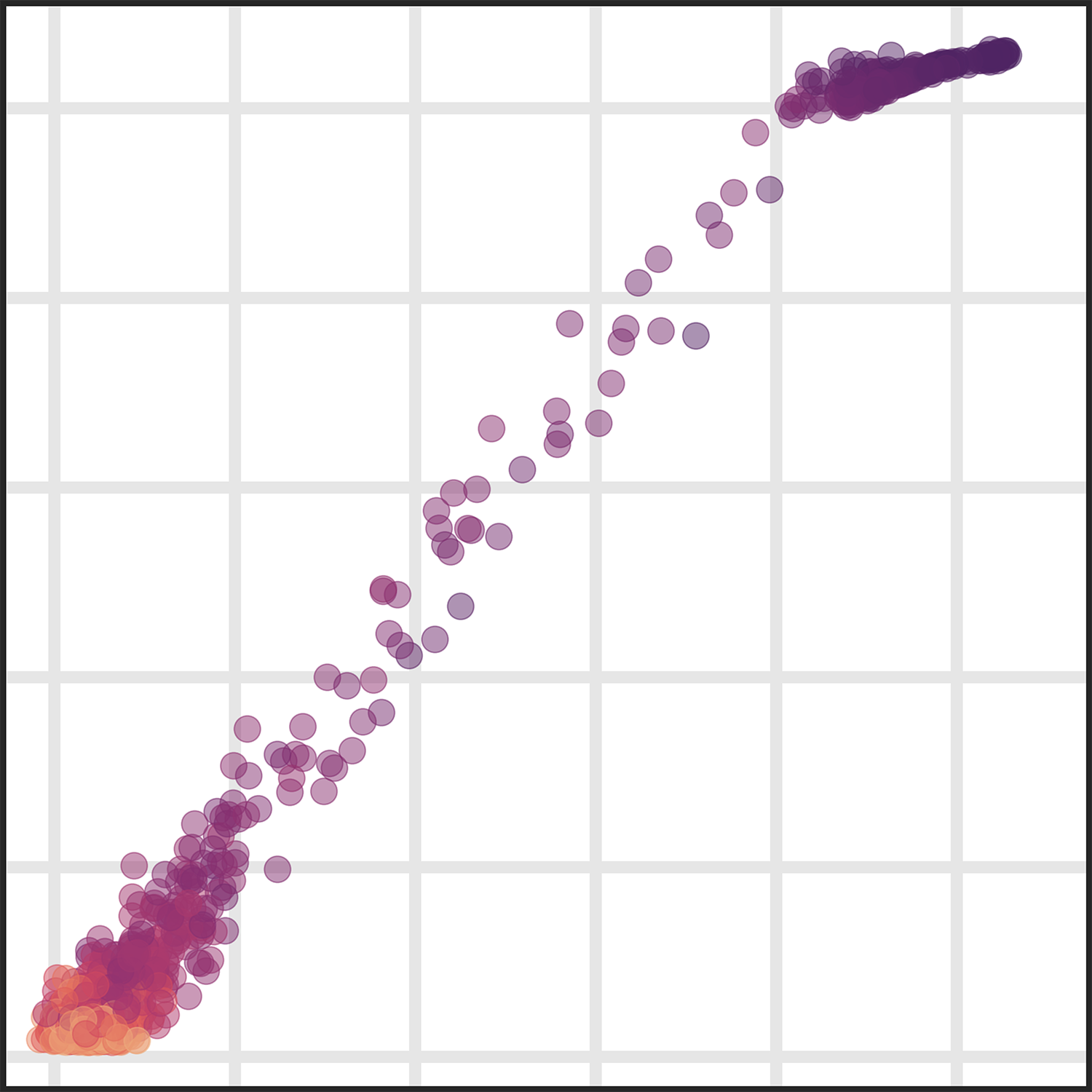};
\end{axis}
\end{tikzpicture}

\begin{tikzpicture}[trim axis right, trim axis left]
\begin{axis}[
    width=0.5\textwidth,
    height=0.015\textwidth,
    xlabel={Standard deviation $\ell$ of the noise added to the expert},
    xtick={0, 0.3, 0.6, 0.9, 1.2, 1.5},
    xticklabels={0, 0.3, 0.6, 0.9, 1.2, 1.5},
    ytick={},
    yticklabels={},
]
\addplot graphics [
ymin=0, ymax=1,
xmin=0, xmax=1.5,
]{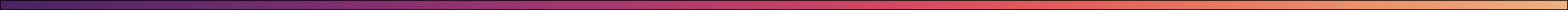};
\end{axis}
\end{tikzpicture}
\vspace{-4pt}

\caption{Calibration plot comparing the proxy reward with the original reward function of the benchmark domains. Each point represents the average of the sum of each reward function, over 5 trajectories. Trajectories are generated by adding noise $\mathcal{N}(0,\ell^2)$ to the expert policy. 
The calibration plots show a strong correlation between the proxy reward and the true task reward.} \label{fig:calib}
\vspace{-8pt}
\end{figure*}

\vspace{-12pt}
\subsection{Analysis and Ablations}
\label{sec:analysis}

To better understand the performance of our approach, in this section, we perform additional analysis to test the quality and importance of various components. These results fill the gap in knowledge left by previous work leveraging the Wasserstein distance in IL, examining hyperparameters such as the regularization parameter in the Sinkhorn distance or the effect of using different distance metrics, and provide direct comparison between the various approximations available to use when comparing policy trajectories.

\begin{table*}[b]
\centering
\setlength{\tabcolsep}{2pt}
\footnotesize	
\begin{tabularx}{0.99\textwidth}{l*{15}{>{\raggedleft\arraybackslash}X}}
\toprule
Environment &  \multicolumn{3}{c}{Hopper}{} &
  \multicolumn{3}{c}{Walker2d}{} &
  \multicolumn{3}{c}{HalfCheetah}{} 
  & \multicolumn{3}{c}{Ant}{} 
  & \multicolumn{3}{c}{Humanoid}{}\\
Space & (s) & (s, s') & (s, a) & (s) & (s, s') & (s, a) & (s) & (s, s') & (s, a) & (s) & (s, s') & (s, a) & (s) & (s, s') & (s, a) \\
  \midrule
{OPOLO} 
& 5.91 & 8.40 & 6.33 & 3.02 & 4.32 & 3.47 & \textbf{1.60} & \textbf{2.39} & \textcolor{blue}{\textbf{1.91}} & 4.64 & 7.24 & \textcolor{blue}{\textbf{5.05}} & 80.75 & 114.53 & 81.90 \\
{PWIL -- $(s)$} 
& 1.74 & 2.56 & 2.38 & \textbf{2.04} & \textbf{2.96} & \textcolor{blue}{\textbf{2.78}} & 6.48 & 9.27 & 6.93 & \textbf{3.83} & 6.00 & 5.90 & 53.52 & 76.06 & 54.94 \\
{OOPS+TD3}    
& \textbf{1.66} & \textbf{2.38} & \textcolor{blue}{\textbf{2.06}} & 2.28 & 3.27 & 3.02 & \textbf{1.63} & \textbf{2.41} & \textcolor{blue}{\textbf{2.01}} & \textbf{3.83} & \textbf{5.90} & \textcolor{blue}{\textbf{5.17}} & \textbf{25.64} & \textbf{37.03} & \textcolor{blue}{\textbf{27.63}} \\
\bottomrule
\end{tabularx}
\caption{Final Wasserstein distance in state occupancy, state transition, and state-action space of the 10 final trained agent rollouts to the expert trajectories for different ILfO algorithms, lower is better. We highlight in \textcolor{blue}{blue} the best performing agent in state-action space, considered ground truth in this experiment, and \textbf{bold} the best performing agent according to each metric. Agents were trained using 10 expert demonstration trajectories, for 1M timesteps. Distances are averaged over 10 reference expert trajectories.} \label{table:wasserstein_comparison} 
\vspace{-12pt}
\end{table*}
\textbf{Accuracy of proxy reward.} OOPS generates a proxy reward function that minimizes the Wasserstein distance between the learner's trajectories and the demonstrated expert trajectories. We evaluate the correlation between this proxy reward and the true environment reward. To do so, we 
collect a dataset of varied trajectory quality using the expert policy from the main results, with added Gaussian noise $\mathcal{N}(0, \ell^2)$ with $\ell \in [0,1.5]$. \autoref{fig:calib} shows the calibration plots between the proxy reward and the original task reward, showing a strong correlation in every environment. 

Next, we compare the quality of trajectories in terms of the Wasserstein distance rather than the true environment reward. In \autoref{table:wasserstein_comparison}, we compare the Wasserstein distance between the expert trajectories and the final policy rollouts obtained at the end of training from each of the \mbox{top-3} performing methods (OOPS, OPOLO, PWIL--$(s)$). 
The Wasserstein distance is measured in three spaces: state-only~$(s)$, state-transition~$(s,s')$, and state-action~$(s,a)$.

We find that OOPS obtains the lowest state-action Wasserstein distance to the expert trajectories in four of the five studied environments, with Walker2d being the only disagreement with the previous experiment, as even though OOPS+TD3 obtains a better task reward in \autoref{table:results}, \mbox{PWIL--$(s)$} obtains a lower state-action Wasserstein distance to the expert. 

Finally, to further evaluate the quality of the Wasserstein distance used by PWIL, we take OOPS and replace the Sinkhorn algorithm with the greedy formulation $W_{greedy}$ proposed by PWIL to compute the Wasserstein distance in $(s,s')$ space. The results are reported in \autoref{table:ablat2} (under $W_{greedy}$), and show a loss in performance. 

\begin{wrapfigure}{r}{0.6\textwidth}
\centering
\vspace{-8pt}

\begin{minipage}{0.75\linewidth}
\begin{tikzpicture}[trim axis right, trim axis left]
\begin{axis}[
    width=\textwidth,
    height=0.08\textwidth,
    xtick={2.8, 3.0, 3.2, 3.4, 3.6, 3.8},
    xticklabels={2.8, 3.0, 3.2, 3.4, 3.6, 3.8},
    ytick={},
    yticklabels={},
]
\addplot graphics [
ymin=-174.62, ymax=3787.98,
xmin=2.72, xmax=3.92,
]{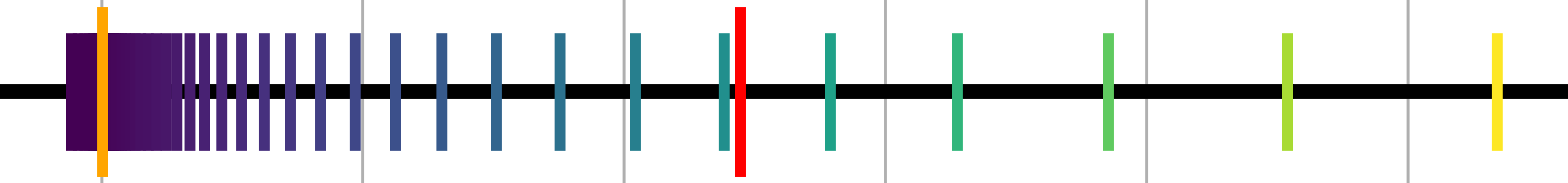};
\end{axis}
\end{tikzpicture}    
\end{minipage}
\vspace{-4pt}

\vspace{-4pt}
\begin{center}
\fcolorbox{gray}{white}{

\begin{minipage}{0.25\textwidth}
\begin{tikzpicture}[trim axis right]
\begin{axis}[
    width=1\textwidth,
    height=0.06\textwidth,
    xlabel={Sinkhorn hyperparameter $\lambda$},
    xtick={0.001, 0.5, 1.0},
    xticklabels={0.001, 0.5, 1.0},
    ytick={},
    yticklabels={},
]
\addplot graphics [
ymin=0, ymax=1,
xmin=0.001, xmax=1,
]{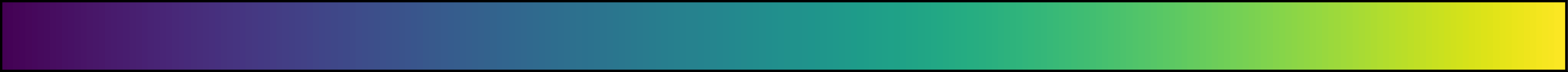};
\end{axis}
\end{tikzpicture}    
\end{minipage}
\hspace{24pt}
\begin{minipage}{0.19\textwidth}
\small \raisebox{-0.3 em}{\mycline{orange}} $W_{\text{simplex}}$ ~ \raisebox{-0.3 em}{\mycline{red}} $W_{\text{greedy}}$
\vspace{8pt}
\end{minipage}

}
\end{center}
\vspace{-4pt}

\caption{Wasserstein distances between the 10 final rollout trajectories of OOPS+TD3 and the expert on the Hopper environment, using different solvers for the coupling matrix~$P$ ($W_{\text{greedy}}$ and $W_{\text{simplex}}$) compared against the Sinkhorn distance $W_{\text{Sk}}$ when varying the parameter $\lambda$. Results are averaged over 10 expert trajectories. The Sinkhorn distance, for low enough values of $\lambda$ computes a tighter upper bound to the Wasserstein distance estimates than $W_{\text{greedy}}$~\citep{PWIL}. Results for the other environments can be found in the Appendix.} \label{fig:approx_comparison}
\vspace{-11pt}
\end{wrapfigure}

\textbf{Quality of estimated Wasserstein distance.} 
In \autoref{fig:approx_comparison}, we compare the quality of different approximations of the state transition Wasserstein distance: the Sinkhorn distance $W_{\text{Sk}}$ with varying $\lambda$, the network simplex solver $W_{\text{simplex}}$ introduced in \citep{emdsolver}, and $W_{\text{greedy}}$ proposed for PWIL~\citep{PWIL}. Additional results can be found in the Appendix.

To compare each approach, we compute the Wasserstein distance between trajectories generated by the final policy of OOPS+TD3 and the expert trajectories, using each of the various approximations. 
Each method results in different estimates of the coupling matrix~$P$; they provide an upper bound on the true Wasserstein distance, where lower estimates of the Wasserstein distance are a tighter bound. We find that for very low values of $\lambda$, $W_{\text{Sk}}$ computes lower cost couplings than $W_{\text{simplex}}$, and up to $\lambda \approx 0.4$ obtains 
better approximations than $W_{\text{greedy}}$. 

\begin{wraptable}{r}{0.57\textwidth}
\vspace{-16pt}
\centering
\setlength{\tabcolsep}{2pt}
\newcommand{\pp}{\phantom{-}}
\footnotesize
\begin{tabularx}{0.57\textwidth}{l*5{>{\centering\arraybackslash}X}}
\toprule
& Hopper & Walker2d & HalfCheetah & Ant & Humanoid \\
\midrule
\multicolumn{6}{l}{\quad Occupancy (Default: $(s,s')$)} \\ 
\midrule
State only               & \po\pp0.10 & -33.93 & \po-0.57 & \po\pp0.30 & \po-0.94 \\
\midrule
\multicolumn{6}{l}{\quad Wasserstein Distance Solver (Default: $\lambda=0.05$)} \\ 
\midrule
$W_{\text{greedy}}$ & -14.99 & \po-7.75 & -45.46 & \po-0.79 & -19.21 \\
$W_{\text{simplex}}$      &     -10.91     &    \po-6.09    &    \po-1.03     &    \po-2.40      &  -33.99     \\
$\lambda=0.005$      &     \po-3.12      &  \po-2.65     &    \po-0.35       &    \po-1.48       &  \po-2.34      \\
$\lambda=0.1$      &     \po-1.72      &    \po-3.87     &   \po-1.39       &   \po-3.88       &   \po-9.18      \\
$\lambda=0.5$      &     -58.64      &    -25.20     &    -15.09       &    -10.95      &  -24.44      \\
\midrule
\multicolumn{6}{l}{\quad Distance Metric (Default: $W_1, d=\sqrt{||\cdot||_2}$)} \\ 
\midrule
$W_2, d=||\cdot||_2$      &     -36.52      &    -21.83    &    -11.88       &    -16.10     &  -47.92      \\
$W_1, d=||\cdot||_2$      &     \po-4.61      &    \po-1.42    &    \po-1.73      &    \po-1.95     &   -22.80      \\
$W_1, d=\cos$      &     \po\pp0.13      &   -10.04    &    \po-4.09      &    \po-2.84     &  -34.63     \\
\midrule
\multicolumn{6}{l}{\quad Adversarial Distance (Default: Unused)} \\ 
\midrule
SIL -- $(s,s')$   &   -82.50    &   -91.61   &   -119.10   &     -124.88   &    -90.83   \\
$\text{OOPS}_{\text{adv}}$     &    -21.09    &    -76.58  &  -101.84       &     \po-17.68    &   -97.87    \\

\bottomrule
\end{tabularx}
 \caption{Results for different variations of OOPS in terms of \emph{percent difference}. All results use 10 expert trajectories and are averaged across 5 seeds and 10 evaluations. 
 State only uses $W_1$ over $(s)$ rather than $(s,s')$. Wasserstein distance solver modifies the solver used by OOPS to determine the coupling matrix $P$. Adversarial distance refers to the use of the adversarial distance function from SIL~\citep{SIL} and also includes the full SIL method for comparison.} 
 \label{table:ablat2}
\vspace{-16pt}
\end{wraptable} 
Next, we compare these three approaches for computing the Wasserstein distance in terms of performance. The results are shown in \autoref{table:ablat2} (Wasserstein Distance Solver). 
Unsurprisingly, large values of $\lambda$, which approximate the Wasserstein distance~$W_1$ poorly, 
results in lower performance. For sufficiently small values of $\lambda$, we find that OOPS+TD3 maintains a consistent performance. This suggests that $\lambda$ can generally be ignored and left to a default value. 

Finally, we attempt different settings for the Wasserstein distance. In \autoref{table:ablat2} we display the change in performance from OOPS when using $W_1$ or $W_2$ when the distance metric $d$ is the Euclidean distance $||\cdot||_2$, and $W_1$ when $d$ is the cosine distance. OOPS uses $W_1$ with the square root of the Euclidean distance, which de-emphasizes large differences in magnitude in a similar fashion to the cosine distance. We find that this choice of $d$ provides significant benefits in high dimensional domains (Humanoid) where magnitudes matter but can vary significantly. We also compare with the learned adversarial distance metric used by SIL~\citep{SIL} (denoted $\text{OOPS}_\text{adv}$) and find that while this version outperforms vanilla SIL, the adversarial component is harmful.

\textbf{Transition vs.\ state occupancy.} For OOPS, we define trajectories by their state-next-state transitions~$(s,s')$, rather than individual states~$s$. Matching based on states can potentially admit multiple minimums since trajectories with the same states out of order can still minimize the state occupation distributional distance. Furthermore, if the reward function is based on state \textit{and} action, then it is clear that only matching state occupancy is insufficient. 
Since expert actions are unavailable in the ILfO setting, we must rely on $(s,s')$.  
We posit that enforcing a local ordering of states provides a higher fidelity signal for ILfO. We validate this empirically in our ablations (\autoref{table:ablat2} - Occupancy). While using state-only occupancy matches the performance of OOPS+TD3 in most environments, there is a large drop in performance in Walker2d. This aligns with our intuition: matching by state occupancy will often work but can be problematic in certain environments depending on the state representation and transition dynamics.

\section{Conclusion}

In this paper, we introduce OOPS, an ILfO algorithm that produces a reward function that minimizes the Wasserstein distance between the state transition trajectory of the expert and the imitation agent. We validate our approach through extensive experiments and demonstrate that OOPS surpasses the current state-of-the-art methods in the ILfO setting across benchmark and robotics domains. Combined with off-policy RL, OOPS exhibits exceptional sample efficiency and low variance in performance, key qualities for the practical deployment of IL algorithms on real systems.

\bibliography{main}

\begin{thebibliography}{42}
\providecommand{\natexlab}[1]{#1}
\providecommand{\url}[1]{\texttt{#1}}
\expandafter\ifx\csname urlstyle\endcsname\relax
  \providecommand{\doi}[1]{doi: #1}\else
  \providecommand{\doi}{doi: \begingroup \urlstyle{rm}\Url}\fi

\bibitem[Arjovsky \& Bottou(2017)Arjovsky and Bottou]{arjovsky2017towards}
Martin Arjovsky and Leon Bottou.
\newblock Towards principled methods for training generative adversarial networks.
\newblock In \emph{International Conference on Learning Representations}, 2017.
\newblock URL \url{https://openreview.net/forum?id=Hk4_qw5xe}.

\bibitem[Arjovsky et~al.(2017)Arjovsky, Chintala, and Bottou]{wgan}
Martin Arjovsky, Soumith Chintala, and L{\'e}on Bottou.
\newblock Wasserstein generative adversarial networks.
\newblock In \emph{International conference on machine learning}, pp.\  214--223. PMLR, 2017.

\bibitem[Bonneel et~al.(2011)Bonneel, van~de Panne, Paris, and Heidrich]{emdsolver}
Nicolas Bonneel, Michiel van~de Panne, Sylvain Paris, and Wolfgang Heidrich.
\newblock Displacement interpolation using lagrangian mass transport.
\newblock \emph{ACM Trans. Graph.}, 30\penalty0 (6):\penalty0 1–12, dec 2011.
\newblock ISSN 0730-0301.
\newblock \doi{10.1145/2070781.2024192}.
\newblock URL \url{https://doi.org/10.1145/2070781.2024192}.

\bibitem[Bonneel et~al.(2015)Bonneel, Rabin, Peyr{\'e}, and Pfister]{bonneel2015sliced}
Nicolas Bonneel, Julien Rabin, Gabriel Peyr{\'e}, and Hanspeter Pfister.
\newblock Sliced and radon wasserstein barycenters of measures.
\newblock \emph{Journal of Mathematical Imaging and Vision}, 51\penalty0 (1):\penalty0 22--45, 2015.

\bibitem[Brockman et~al.(2016)Brockman, Cheung, Pettersson, Schneider, Schulman, Tang, and Zaremba]{OpenAIGym}
Greg Brockman, Vicki Cheung, Ludwig Pettersson, Jonas Schneider, John Schulman, Jie Tang, and Wojciech Zaremba.
\newblock Openai gym, 2016.

\bibitem[Chang et~al.(2022)Chang, Higuera, Fujimoto, Meger, and Dudek]{chang2022flow}
Wei-Di Chang, Juan Camilo~Gamboa Higuera, Scott Fujimoto, David Meger, and Gregory Dudek.
\newblock Il-flow: Imitation learning from observation using normalizing flows.
\newblock \emph{arXiv preprint arXiv:2205.09251}, 2022.

\bibitem[Coumans \& Bai(2016)Coumans and Bai]{coumans2016pybullet}
Erwin Coumans and Yunfei Bai.
\newblock Pybullet, a python module for physics simulation for games, robotics and machine learning, 2016.

\bibitem[Cuturi(2013)]{cuturi2013sinkhorn}
Marco Cuturi.
\newblock Sinkhorn distances: Lightspeed computation of optimal transport.
\newblock \emph{Advances in neural information processing systems}, 26, 2013.

\bibitem[Dadashi et~al.(2020)Dadashi, Hussenot, Geist, and Pietquin]{PWIL}
Robert Dadashi, Leonard Hussenot, Matthieu Geist, and Olivier Pietquin.
\newblock Primal wasserstein imitation learning.
\newblock In \emph{International Conference on Learning Representations}, 2020.

\bibitem[Desai et~al.(2020)Desai, Durugkar, Karnan, Warnell, Hanna, Stone, and Sony]{desai2020imitation}
Siddharth Desai, Ishan Durugkar, Haresh Karnan, Garrett Warnell, Josiah Hanna, Peter Stone, and AI~Sony.
\newblock An imitation from observation approach to sim-to-real transfer.
\newblock 2020.

\bibitem[Durugkar et~al.(2021)Durugkar, Tec, Niekum, and Stone]{durugkar2021adversarial}
Ishan Durugkar, Mauricio Tec, Scott Niekum, and Peter Stone.
\newblock Adversarial intrinsic motivation for reinforcement learning.
\newblock \emph{Advances in Neural Information Processing Systems}, 34:\penalty0 8622--8636, 2021.

\bibitem[Fu et~al.(2017)Fu, Luo, and Levine]{fu2017learning}
Justin Fu, Katie Luo, and Sergey Levine.
\newblock Learning robust rewards with adversarial inverse reinforcement learning.
\newblock \emph{arXiv preprint arXiv:1710.11248}, 2017.

\bibitem[Fujimoto et~al.(2018)Fujimoto, van Hoof, and Meger]{fujimoto2018addressing}
Scott Fujimoto, Herke van Hoof, and David Meger.
\newblock Addressing function approximation error in actor-critic methods.
\newblock In \emph{International Conference on Machine Learning}, volume~80, pp.\  1587--1596. PMLR, 2018.

\bibitem[Fujimoto et~al.(2020)Fujimoto, Meger, and Precup]{fujimoto2020equivalence}
Scott Fujimoto, David Meger, and Doina Precup.
\newblock An equivalence between loss functions and non-uniform sampling in experience replay.
\newblock \emph{Advances in neural information processing systems}, 33:\penalty0 14219--14230, 2020.

\bibitem[Ghasemipour et~al.(2020)Ghasemipour, Zemel, and Gu]{ghasemipour2020divergence}
Seyed Kamyar~Seyed Ghasemipour, Richard Zemel, and Shixiang Gu.
\newblock A divergence minimization perspective on imitation learning methods.
\newblock In \emph{Conference on Robot Learning}, pp.\  1259--1277. PMLR, 2020.

\bibitem[Haarnoja et~al.(2018{\natexlab{a}})Haarnoja, Zhou, Abbeel, and Levine]{haarnoja2018soft}
Tuomas Haarnoja, Aurick Zhou, Pieter Abbeel, and Sergey Levine.
\newblock Soft actor-critic: Off-policy maximum entropy deep reinforcement learning with a stochastic actor.
\newblock In \emph{International Conference on Machine Learning}, volume~80, pp.\  1861--1870. PMLR, 2018{\natexlab{a}}.

\bibitem[Haarnoja et~al.(2018{\natexlab{b}})Haarnoja, Zhou, Hartikainen, Tucker, Ha, Tan, Kumar, Zhu, Gupta, Abbeel, et~al.]{haarnoja2018applications}
Tuomas Haarnoja, Aurick Zhou, Kristian Hartikainen, George Tucker, Sehoon Ha, Jie Tan, Vikash Kumar, Henry Zhu, Abhishek Gupta, Pieter Abbeel, et~al.
\newblock Soft actor-critic algorithms and applications.
\newblock \emph{arXiv preprint arXiv:1812.05905}, 2018{\natexlab{b}}.

\bibitem[Haldar et~al.(2023)Haldar, Mathur, Yarats, and Pinto]{haldar2023watch}
Siddhant Haldar, Vaibhav Mathur, Denis Yarats, and Lerrel Pinto.
\newblock Watch and match: Supercharging imitation with regularized optimal transport.
\newblock In \emph{Conference on Robot Learning}, pp.\  32--43. PMLR, 2023.

\bibitem[Ho \& Ermon(2016)Ho and Ermon]{ho2016generative}
Jonathan Ho and Stefano Ermon.
\newblock Generative adversarial imitation learning.
\newblock In \emph{Advances in Neural Information Processing Systems}, pp.\  4565--4573, 2016.

\bibitem[Jaegle et~al.(2021)Jaegle, Sulsky, Ahuja, Bruce, Fergus, and Wayne]{jaegle2021imitation}
Andrew Jaegle, Yury Sulsky, Arun Ahuja, Jake Bruce, Rob Fergus, and Greg Wayne.
\newblock Imitation by predicting observations.
\newblock In \emph{International Conference on Machine Learning}, pp.\  4665--4676. PMLR, 2021.

\bibitem[Kalashnikov et~al.(2018)Kalashnikov, Irpan, Pastor, Ibarz, Herzog, Jang, Quillen, Holly, Kalakrishnan, Vanhoucke, et~al.]{kalashnikov2018scalable}
Dmitry Kalashnikov, Alex Irpan, Peter Pastor, Julian Ibarz, Alexander Herzog, Eric Jang, Deirdre Quillen, Ethan Holly, Mrinal Kalakrishnan, Vincent Vanhoucke, et~al.
\newblock Scalable deep reinforcement learning for vision-based robotic manipulation.
\newblock In \emph{Conference on Robot Learning}, pp.\  651--673, 2018.

\bibitem[Ke et~al.(2019)Ke, Barnes, Sun, Lee, Choudhury, and Srinivasa]{ke2019imitation}
Liyiming Ke, Matt Barnes, Wen Sun, Gilwoo Lee, Sanjiban Choudhury, and Siddhartha Srinivasa.
\newblock Imitation learning as $ f $-divergence minimization.
\newblock \emph{arXiv preprint arXiv:1905.12888}, 2019.

\bibitem[Kidambi et~al.(2021)Kidambi, Chang, and Sun]{kidambi2021mobile}
Rahul Kidambi, Jonathan Chang, and Wen Sun.
\newblock Mobile: Model-based imitation learning from observation alone.
\newblock \emph{Advances in Neural Information Processing Systems}, 34, 2021.

\bibitem[Kostrikov et~al.(2018)Kostrikov, Agrawal, Dwibedi, Levine, and Tompson]{kostrikov2018discriminator}
Ilya Kostrikov, Kumar~Krishna Agrawal, Debidatta Dwibedi, Sergey Levine, and Jonathan Tompson.
\newblock Discriminator-actor-critic: Addressing sample inefficiency and reward bias in adversarial imitation learning.
\newblock In \emph{International Conference on Learning Representations}, 2018.

\bibitem[Kostrikov et~al.(2019)Kostrikov, Agrawal, Dwibedi, Levine, and Tompson]{DACkostrikov}
Ilya Kostrikov, Kumar~Krishna Agrawal, Debidatta Dwibedi, Sergey Levine, and Jonathan Tompson.
\newblock Discriminator-actor-critic: Addressing sample inefficiency and reward bias in adversarial imitation learning.
\newblock In \emph{7th International Conference on Learning Representations, {ICLR} 2019, New Orleans, LA, USA, May 6-9, 2019}. OpenReview.net, 2019.
\newblock URL \url{https://openreview.net/forum?id=Hk4fpoA5Km}.

\bibitem[Lillicrap et~al.(2015)Lillicrap, Hunt, Pritzel, Heess, Erez, Tassa, Silver, and Wierstra]{DDPG}
Timothy~P Lillicrap, Jonathan~J Hunt, Alexander Pritzel, Nicolas Heess, Tom Erez, Yuval Tassa, David Silver, and Daan Wierstra.
\newblock Continuous control with deep reinforcement learning.
\newblock \emph{arXiv preprint arXiv:1509.02971}, 2015.

\bibitem[Ni et~al.(2020)Ni, Sikchi, Wang, Gupta, Lee, and Eysenbach]{firl2020corl}
Tianwei Ni, Harshit Sikchi, Yufei Wang, Tejus Gupta, Lisa Lee, and Ben Eysenbach.
\newblock f-irl: Inverse reinforcement learning via state marginal matching.
\newblock In \emph{Conference on Robot Learning}, 2020.

\bibitem[Papagiannis \& Li(2020)Papagiannis and Li]{SIL}
Georgios Papagiannis and Yunpeng Li.
\newblock Imitation learning with sinkhorn distances.
\newblock \emph{arXiv preprint arXiv:2008.09167}, 2020.

\bibitem[Peyr{\'e} et~al.(2019)Peyr{\'e}, Cuturi, et~al.]{peyre2019computational}
Gabriel Peyr{\'e}, Marco Cuturi, et~al.
\newblock Computational optimal transport: With applications to data science.
\newblock \emph{Foundations and Trends{\textregistered} in Machine Learning}, 11\penalty0 (5-6):\penalty0 355--607, 2019.

\bibitem[Schulman et~al.(2015)Schulman, Levine, Abbeel, Jordan, and Moritz]{trpo}
John Schulman, Sergey Levine, Pieter Abbeel, Michael Jordan, and Philipp Moritz.
\newblock Trust region policy optimization.
\newblock In \emph{International Conference on Machine Learning}, pp.\  1889--1897, 2015.

\bibitem[Sinkhorn(1967)]{sinkhorn1967diagonal}
Richard Sinkhorn.
\newblock Diagonal equivalence to matrices with prescribed row and column sums.
\newblock \emph{The American Mathematical Monthly}, 74\penalty0 (4):\penalty0 402--405, 1967.

\bibitem[Stanczuk et~al.(2021)Stanczuk, Etmann, Kreusser, and Sch{\"o}nlieb]{stanczuk2021wasserstein}
Jan Stanczuk, Christian Etmann, Lisa~Maria Kreusser, and Carola-Bibiane Sch{\"o}nlieb.
\newblock Wasserstein gans work because they fail (to approximate the wasserstein distance).
\newblock \emph{arXiv preprint arXiv:2103.01678}, 2021.

\bibitem[Sun et~al.(2019)Sun, Vemula, Boots, and Bagnell]{sun2019provably}
Wen Sun, Anirudh Vemula, Byron Boots, and Drew Bagnell.
\newblock Provably efficient imitation learning from observation alone.
\newblock In \emph{International Conference on Machine Learning}, pp.\  6036--6045. PMLR, 2019.

\bibitem[Tan et~al.(2018)Tan, Zhang, Coumans, Iscen, Bai, Hafner, Bohez, and Vanhoucke]{Tan-RSS-18}
Jie Tan, Tingnan Zhang, Erwin Coumans, Atil Iscen, Yunfei Bai, Danijar Hafner, Steven Bohez, and Vincent Vanhoucke.
\newblock Sim-to-real: Learning agile locomotion for quadruped robots.
\newblock In \emph{Proceedings of Robotics: Science and Systems}, Pittsburgh, Pennsylvania, June 2018.
\newblock \doi{10.15607/RSS.2018.XIV.010}.

\bibitem[Todorov et~al.(2012)Todorov, Erez, and Tassa]{mujoco}
Emanuel Todorov, Tom Erez, and Yuval Tassa.
\newblock Mujoco: A physics engine for model-based control.
\newblock In \emph{IEEE/RSJ International Conference on Intelligent Robots and Systems (IROS)}, pp.\  5026--5033. IEEE, 2012.

\bibitem[Torabi et~al.(2018{\natexlab{a}})Torabi, Warnell, and Stone]{torabi2018behavioral}
Faraz Torabi, Garrett Warnell, and Peter Stone.
\newblock Behavioral cloning from observation.
\newblock In \emph{Proceedings of the 27th International Joint Conference on Artificial Intelligence}, pp.\  4950--4957, 2018{\natexlab{a}}.

\bibitem[Torabi et~al.(2018{\natexlab{b}})Torabi, Warnell, and Stone]{torabi2018generative}
Faraz Torabi, Garrett Warnell, and Peter Stone.
\newblock Generative adversarial imitation from observation.
\newblock \emph{arXiv preprint arXiv:1807.06158}, 2018{\natexlab{b}}.

\bibitem[Villani(2009)]{villani2009optimal}
C{\'e}dric Villani.
\newblock \emph{Optimal transport: old and new}, volume 338.
\newblock Springer, 2009.

\bibitem[Xiao et~al.(2019)Xiao, Herman, Wagner, Ziesche, Etesami, and Linh]{xiao2019wasserstein}
Huang Xiao, Michael Herman, Joerg Wagner, Sebastian Ziesche, Jalal Etesami, and Thai~Hong Linh.
\newblock Wasserstein adversarial imitation learning.
\newblock \emph{arXiv preprint arXiv:1906.08113}, 2019.

\bibitem[{Zhang} et~al.(2020){Zhang}, {Wang}, {Ma}, {Xia}, {Yang}, {Li}, and {Li}]{wdail}
M.~{Zhang}, Y.~{Wang}, X.~{Ma}, L.~{Xia}, J.~{Yang}, Z.~{Li}, and X.~{Li}.
\newblock Wasserstein distance guided adversarial imitation learning with reward shape exploration.
\newblock In \emph{2020 IEEE 9th Data Driven Control and Learning Systems Conference (DDCLS)}, pp.\  1165--1170, Nov 2020.
\newblock \doi{10.1109/DDCLS49620.2020.9275169}.

\bibitem[Zhu et~al.(2020)Zhu, Lin, Dai, and Zhou]{opolozhu2020}
Zhuangdi Zhu, Kaixiang Lin, Bo~Dai, and Jiayu Zhou.
\newblock Off-policy imitation learning from observations.
\newblock \emph{Advances in Neural Information Processing Systems}, 33:\penalty0 12402--12413, 2020.

\bibitem[Ziebart et~al.(2008)Ziebart, Maas, Bagnell, and Dey]{ziebart2008maximum}
Brian~D Ziebart, Andrew~L Maas, J~Andrew Bagnell, and Anind~K Dey.
\newblock Maximum entropy inverse reinforcement learning.
\newblock In \emph{AAAI}, volume~8, pp.\  1433--1438, 2008.

\end{thebibliography}
\bibliographystyle{rlc}
\newpage
\appendix

\section{Additional Results and Experiments}
\label{sec:appendix1}
\subsection{Comparing Solvers for the State Transition Wasserstein Distance}
\begin{figure}[h]
\raggedright

\begin{minipage}{0.2\linewidth}
\hspace{0pt}
\vfill
{\footnotesize Hopper}
\vspace{6pt}
\vfill
\hspace{0pt}
\end{minipage}%
\begin{minipage}{0.75\linewidth}
\begin{tikzpicture}[trim axis right, trim axis left]
\begin{axis}[
    width=\textwidth,
    height=0.08\textwidth,
    xtick={2.8, 3.0, 3.2, 3.4, 3.6, 3.8},
    xticklabels={2.8, 3.0, 3.2, 3.4, 3.6, 3.8},
    ytick={},
    yticklabels={},
]
\addplot graphics [
ymin=-174.62, ymax=3787.98,
xmin=2.72, xmax=3.92,
]{figures/new_lambdas/Hopper.png};
\end{axis}
\end{tikzpicture}    
\end{minipage}
\vspace{-4pt}

\begin{minipage}{0.2\linewidth}
\hspace{0pt}
\vfill
{\footnotesize Walker2d\vphantom{p}}
\vspace{6pt}
\vfill
\hspace{0pt}
\end{minipage}%
\begin{minipage}{0.75\linewidth}
\begin{tikzpicture}[trim axis right, trim axis left]
\begin{axis}[
    width=\textwidth,
    height=0.08\textwidth,
    xtick={2.8, 3.0, 3.2, 3.4, 3.6, 3.8, 4.0},
    xticklabels={2.8, 3.0, 3.2, 3.4, 3.6, 3.8, 4.0},
    ytick={},
    yticklabels={},
]
\addplot graphics [
ymin=-174.62, ymax=3787.98,
xmin=2.79, xmax=4.17,
]{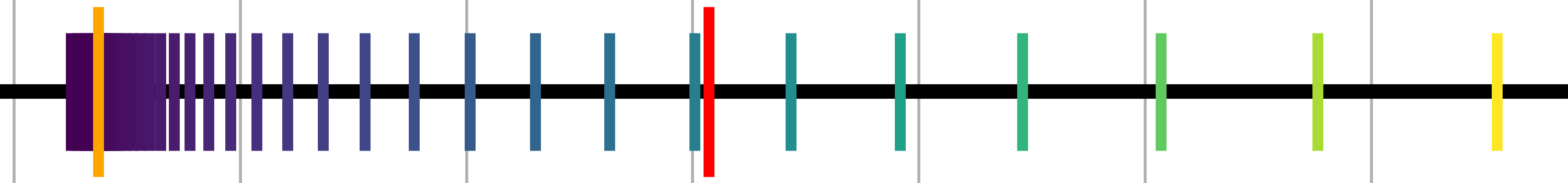};
\end{axis}
\end{tikzpicture}    
\end{minipage}
\vspace{-4pt}

\begin{minipage}{0.2\linewidth}
\hspace{0pt}
\vfill
{\footnotesize HalfCheetah\vphantom{p}}
\vspace{6pt}
\vfill
\hspace{0pt}
\end{minipage}%
\begin{minipage}{0.75\linewidth}
\begin{tikzpicture}[trim axis right, trim axis left]
\begin{axis}[
    width=\textwidth,
    height=0.08\textwidth,
    xtick={2.0, 2.2, 2.4, 2.6, 2.8, 3.0, 3.2, 3.4, 3.6, 3.8, 4.0},
    xticklabels={2.0, 2.2, 2.4, 2.6, 2.8, 3.0, 3.2, 3.4, 3.6, 3.8, 4.0},
    ytick={},
    yticklabels={},
]
\addplot graphics [
ymin=-174.62, ymax=3787.98,
xmin=2.17, xmax=3.59,
]{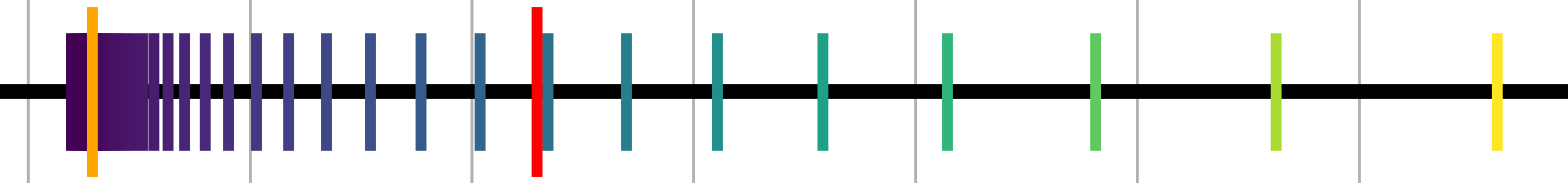};
\end{axis}
\end{tikzpicture}    
\end{minipage}
\vspace{-4pt}

\begin{minipage}{0.2\linewidth}
\hspace{0pt}
\vfill
{\footnotesize Ant\vphantom{p}}
\vspace{6pt}
\vfill
\hspace{0pt}
\end{minipage}%
\begin{minipage}{0.75\linewidth}
\begin{tikzpicture}[trim axis right, trim axis left]
\begin{axis}[
    width=\textwidth,
    height=0.08\textwidth,
    xtick={6.0, 6.2, 6.4, 6.6, 6.8, 7.0, 7.2, 7.4, 7.6},
    xticklabels={6.0, 6.2, 6.4, 6.6, 6.8, 7.0, 7.2, 7.4, 7.6},
    ytick={},
    yticklabels={},
]
\addplot graphics [
ymin=-174.62, ymax=3787.98,
xmin=5.94, xmax=7.64,
]{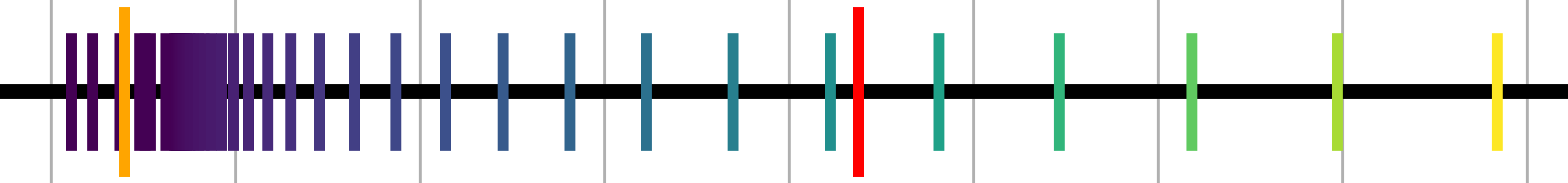};
\end{axis}
\end{tikzpicture}    
\end{minipage}
\vspace{-4pt}

\begin{minipage}{0.2\linewidth}
\hspace{0pt}
\vfill
{\footnotesize Humanoid}
\vspace{16pt}
\vfill
\hspace{0pt}
\end{minipage}%
\begin{minipage}{0.75\linewidth}
\begin{tikzpicture}[trim axis right, trim axis left]
\begin{axis}[
    width=\textwidth,
    height=0.08\textwidth,
    xtick={51.8, 52.0, 52.2, 52.4, 52.6, 52.8, 53.0, 53.2},
    xticklabels={51.8, 52.0, 52.2, 52.4, 52.6, 52.8, 53.0, 53.2},
    ytick={},
    yticklabels={},
    xlabel={Estimated Wasserstein Distance},
]
\addplot graphics [
ymin=-174.62, ymax=3787.98,
xmin=51.70, xmax=53.28,
]{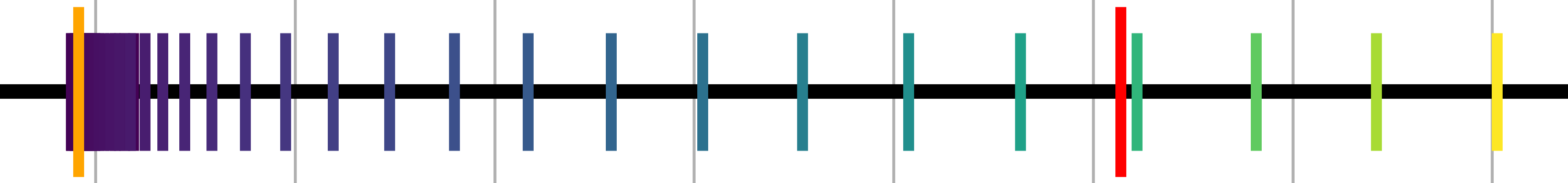};
\end{axis}
\end{tikzpicture}    
\end{minipage}

\vspace{-4pt}
\begin{center}
\fcolorbox{gray}{white}{

\begin{minipage}{0.25\textwidth}
\begin{tikzpicture}[trim axis right]
\begin{axis}[
    width=1\textwidth,
    height=0.06\textwidth,
    xlabel={Sinkhorn hyperparameter $\lambda$},
    xtick={0.001, 0.5, 1.0},
    xticklabels={0.001, 0.5, 1.0},
    ytick={},
    yticklabels={},
]
\addplot graphics [
ymin=0, ymax=1,
xmin=0.001, xmax=1,
]{figures/new_lambdas/better_colorbar_viridis.png};
\end{axis}
\end{tikzpicture}    
\end{minipage}
\hspace{24pt}
\begin{minipage}{0.19\textwidth}
\raggedleft \small \raisebox{-0.3 em}{\mycline{orange}} $W_{\text{simplex}}$ ~ \raisebox{-0.3 em}{\mycline{red}} $W_{\text{greedy}}$
\vspace{8pt}
\end{minipage}

}
\end{center}
\vspace{-4pt}

\caption{Wasserstein distances between the 10 final rollout trajectories of OOPS+TD3 and the expert, using different solvers for the coupling matrix~$P$ ($W_{\text{greedy}}$ and $W_{\text{simplex}}$) compared against the Sinkhorn distance $W_{\text{Sk}}$ when varying the parameter $\lambda$. Results are averaged over 10 expert trajectories. The Sinkhorn distance, for low enough values of $\lambda$ computes a tighter upper bound to the Wasserstein distance estimates than $W_{\text{greedy}}$~\citep{PWIL}.} \label{fig:approx_comparison_full}
\vspace{-12pt}
\end{figure}

We show in \autoref{fig:approx_comparison_full} the full set of results for the comparison of solvers used when computing the Wasserstein distance. See \autoref{sec:analysis} for the description and discussion of this experiment.

\section{Experimental Details}
\label{sec:appendix2}

In \autoref{table:td3_hyperparams}, we list the hyperparameters used for TD3~\citep{fujimoto2018addressing}, our underlying off-policy RL algorithm. On top of these hyperparameters, we use the PAL variant of TD3 for the loss function of the critic~\citep{fujimoto2020equivalence}. In \autoref{table:sk_hyperparams}, we list the hyperparameters for the computation of the Sinkhorn distance~\citep{cuturi2013sinkhorn} used for OOPS across all experiments, except experiments studying the effect of specific hyperparameters (distance metric and $\lambda$). 

\begin{table}[h]
\begin{center}
\begin{tabular}{cc} 
\toprule
    {Parameter}  & {Value}  \\ \midrule
    {$\tau$}     & 3e-3  \\
    {Exploration noise}     & 2e-1  \\
    {Policy noise}     & 1e-1 \\
    {Actor network architecture (hidden)}     & [256]  \\ 
    {Critic network architecture (hidden)}   &  [1024] \\ 
    {Actor LR}  &  3e-4  \\ 
    {Critic LR}    & 3e-4  \\
    {Optimizer}    & Adam  \\
    {Actor non linearity} & ReLU \\
    {Critic non linearity} & ReLU \\
 \bottomrule
\end{tabular}
\end{center}
\caption{TD3 hyperparameters}
 \label{table:td3_hyperparams}
\end{table}

\begin{table}[h]
\begin{center}
\begin{tabular}{cc} 
\toprule
    {Parameter}  & {Value}  \\ \midrule
    {Maximum number of iterations}     & 20000  \\
    {$\lambda$}     & 0.05 \\ 
    {Distance metric}   &  $\sqrt{||\cdot||_2}$\\ 
 \bottomrule
\end{tabular}
\end{center}
\caption{Sinkhorn distance computation hyperparameters}
 \label{table:sk_hyperparams}
\end{table}

\end{document}